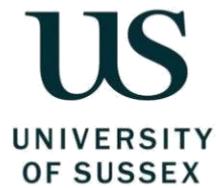

UNIVERSITY
OF SUSSEX

# Autonomous Visual Navigation

*A Biologically Inspired Approach*

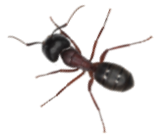

**Sotirios Athanasoulias**
Candidate Number:164677
Supervisor: Dr. Andrew Philippides

# Declaration

This report is submitted as part requirement for the degree of Computer Science & Artificial Intelligence at the University of Sussex. It is the product of my own labour except where indicated in the text. The report may be freely copied and distributed provided the source is acknowledged.

**Signature:**

# Acknowledgments

I would like to express my very great appreciation to Dr Andy Philippides for his valuable advice and continuous support throughout this year. His willingness to give me his time so generously as well as his motivation has been very much appreciated. I have been extremely lucky to have a supervisor who cared so much about my work and who was always available for me. Within this project he helped me to understand the basic principles of a research project and he inspired me to make further investigation into the biologically inspired algorithms. This project could not have been done without him.

# UNIVERSITY OF SUSSEX

SOTIRIOS ATHANASOULIAS

## AUTONOMOUS VISUAL NAVIGATION

### A BIOLOGICALLY INSPIRED APPROACH

## SUMMARY

Inspired by the navigational behaviour observed in the animal kingdom and especially the navigational behaviour of the ants, we attempt to simulate it in an artificial environment by implementing different kinds of biomimetic algorithms. Ants navigate themselves by using retinotopic views and try to move in a position to perceive the world in way to look more like they have memorized it. Using this concept, we implement one robust method, "Perfect Memory", which uses the Snapshot model. Perfect Memory is based on the unrealistic assumption of remembering every single snapshot experienced across a training route. After evaluating the performance of this technique and confirming its robustness, we approach the same problem using Artificial Neural Networks (ANNs) as classifiers. This approach has the advantage of providing a holistic representation of the route and the agent does not need to memorize every single snapshot. The basic idea is that we train an ANN to classify whether a view is part of the route or not using the snapshots as training data. We aim to explore and compare the performance between different ANNs classification techniques using as baseline the Perfect Memory.

## Table of Contents

# 1. Introduction

Autonomous navigation is a field of the Adaptive robotics with increasing interest over the last decades. Navigation, as used in robotics, is the process of monitoring and controlling the movement of an agent along a route (Lee and Kim, 2017). Recent studies show diverse forms of autonomous navigation systems varying from simple wheel robots to unmanned aerial vehicles. The main source of inspiration for implementing such systems is nature and specifically the navigational behavior of the animals. This research project will focus on the visual navigation behaviour of a special case of navigators, the ants, exploring different types of biomimetic algorithms which replicate their navigational behaviour.

Despite their tiny brains, ants perform incredible navigation tasks in complex natural environments (Moller et al., 1998). In order to find their location, ants use a number of different mechanisms such as pheromones and path integration systems. However, for dessert ants the main source of information comes from their learnt visual views (Collett et al., 2006). The main idea of their navigational behavior is that they scan the world and move to the direction that provides the best match with the view they have memorized form a previous traversal. Inspired by this behaviour many engineers, computer scientists and biologists have tried to implement models to replicate it. Some of the most successful and common approaches are those ones who based on the Snapshot model. According to this model, the agent moves to its goal location by comparing its current view with the snapshots of the training route. Within this project we will implement our baseline method, Perfect Memory, which uses Snapshot Model and defines a route in terms of distinct snapshots along a training route. The main disadvantage of this approach is that is based on the unrealistic assumption of remembering every single snapshot experienced across a training route. Another more advanced approach to this problem is the use of artificial neural networks. According to this approach, the agent uses an ANN to train it with the training route images. Its main advantage is that it provides a holistic representation of the route and the agent does not need to remember every single snapshot.

We decided to approach the problem using ANNs as classifiers. Although there have been similar approaches, the innovative part of this research is that we use a simple Multi-Layer Perceptron (MLP). The reason that we decided to use this type of neural network is that MLP is a representative example of a feedforward global neural network and we were already familiar with its structure and the way it works (Zanaty,2012).

The basic idea of this approach is that we train an ANN using both positive (views that are part of the route) and negative (views that are not part of the route) training data to classify whether a view is part of the route or not. Using Artificial Neural Networks approach, reframed the problem of navigation in terms of a familiarity search of views that look similar with the training views (Philippides et al., 2015).

The project focuses on many different aspects regarding the performance of our biomimetic algorithms and tries to analyse what works, what does not work and why. Some interesting observations were that mid-range visual resolution images are better suited for navigation, that the ANNs implementation cannot achieve high performance in corners of the route as well as that it produces better results for smaller number of training data. The main aim of this project is to understand the visual navigation system of the ants, replicate it using "ANNs as classifiers" approach and evaluate its performance using as baseline the Perfect Memory.

The report starts describing the navigational behavior of the ants as well as the previous approaches to this problem. Subsequently, it mentions the baselines methods of the project and finishes by analyzing our ANN implementation and possible improvements of our approach.

## 1.1 Motivation

Insect inspired algorithms and their implementation on visual navigation is the main research field of University of Sussex Brains on Board team. Brains on Board is a collaboration between Sheffield, Sussex and Queen Mary Universities to investigate insect's behavior and build biomimetic algorithms. One of their current project focuses on the visual navigation behavior of the ants using ANNs. Although, a similar approach has been conducted this project extend the approach of using ANNs as classifiers as it focuses on different aspects and also uses different types of classifiers. Finally, this topic is relatively new so there is dearth of research on using ANNs to imitate the navigational behavior of the ants. That means that there is a potential for new observations and ideas which will help the further development of the project.

# 2. Professional and Ethical Considerations

## 2.1 Professional Considerations

This project abides by BCS, the Institute for IT's code of conduct. (Bcs.org,)
Some of the key sections of the code of conduct that they are most relevant to this project are:

**Public interest**

**1. a)** "Have due regard for public health, privacy, security and wellbeing of other and environment." – the project will include a simulated world, an artificial agent and other programming tools so it will not affect members of the public or the environment.

**1. b)** "Have due regard for the legitimate rights of Third Parties." – Any third-party libraries, databases, images or software used will have the proper credit given.

**Professional Competence and Integrity**

**2. c)** "Develop your professional knowledge, skills and competence on a continuing basis, maintaining awareness of technological developments, procedures, and standards that are relevant to your field." – During this project new skills and knowledge relevant to computer science and artificial intelligence will be learnt to the expected standard in order to come up with an outstanding solution to the autonomous visual navigation topic.

**2. e)** "Respect and value alternative viewpoints and, seek, accept and offer honest criticism of work." – During my project I will use different viewpoints on autonomous visual navigation in order to implement different models, evaluate and compare their performances. Also, I will get feedback and alternative viewpoints from my supervisor and his PhD students which will be appreciated and they will help me along the implementation of my project.

**2. f)** "Avoid injuring others, their property, reputation, or employment by false or malicious or negligent action or inaction." – This project involves a simulated world, an artificial agent and other programming tools so there is no possibility to cause injuring to someone or to his/her property. Also this project is about imitating the visual navigation system of the ants so there is no possibility to cause the other risks referenced in 2.f) too.

## 2.2 Ethical Considerations

This project will be implemented and evaluated using a simulated world, an artificial agent and specific programming tools. Therefore, no participants will be involved in my research and as a result no review form for ethical considerations is required.

## 2.3 Ethical considerations of AI / Robotics

This project can be considered as an Artificial Intelligence / Insect inspired robotic project. Brains on Board is a research team which has been formed as a collaboration between Sheffield Sussex and Queen Mary Universities with the aim to make research and implement biomimetic algorithms inspired by the insect's behavior. The team has proposed an ethical statement (Appendix 7.1) regarding their field of research which covers the ethical AI/Robotics considerations of our project as well.

The main consideration of their statement that is also applied in this project is that the "AI agents/Robots are multi-use tools and should be designed, operated as far as are practicable to comply with the existing laws, fundamental rights and freedoms" – The AI agent designed and used within this project will have bio-mimetic behavior based on the ant's brains. All the algorithms will be designed as basic research tools to better understand the autonomy of animals and the AI agents. Finally, our agents will be designed to comply with the existing laws and fundamental rights.

# 3. Background

In this section, we are looking into some background research regarding the visual navigation of the ants as well as some previous approaches that try to imitate this behavior. Firstly, we describe the visual navigation of the ants and state some interesting facts about their behavior. Subsequently, we focus on the Snapshot model, which is one of the most common approaches that they have been used to replicate the navigational system of the insects. Finally, we describe some more advanced approaches that use neural networks to solve the same problem. All these components were the basic inspiration for both my "perfect memory" and ANN as classifier implementations.

## 3.1 Navigation in ants

Many animals and insects of the animal kingdom use the visual information that they get from their environment for navigation. One interesting case of visual navigation that this project focuses on is the visual navigation system of the ants. Ants and especially the dessert ants, where the extreme heat evaporates their pheromone trails, are good visual navigators as they use visual information to guide foraging routes between their nest and a stable food site. An interesting fact is that they are so capable that they can memorize a long complex route from the first time they traverse it (Philippides et.al, 2012).

As expert navigators, ants are always able to return to their nests by using a basic dead reckoning strategy (Wehner and Srinivasan, 2003). Specifically, they use celestial information as a compass (Wehner and Wehner, 1986) as well as a step-counting method for odometry (Wittlinger, 2006) and they continuously calculate all the essential information to take a direct path back to their home at any time. This ability allows ants to safely explore the world and obtain navigationally useful visual information.

According to many experiments that have been conducted to many different insects, we have concluded that they store all these visual information as egocentric views of their environment as seen from a goal location or from correct route directions. This navigational strategy is referred to as "view-based matching" (Graham and Philippides, 2017). Subsequently, they recall all these stored views, compare them with their current view and navigate towards to their goal.

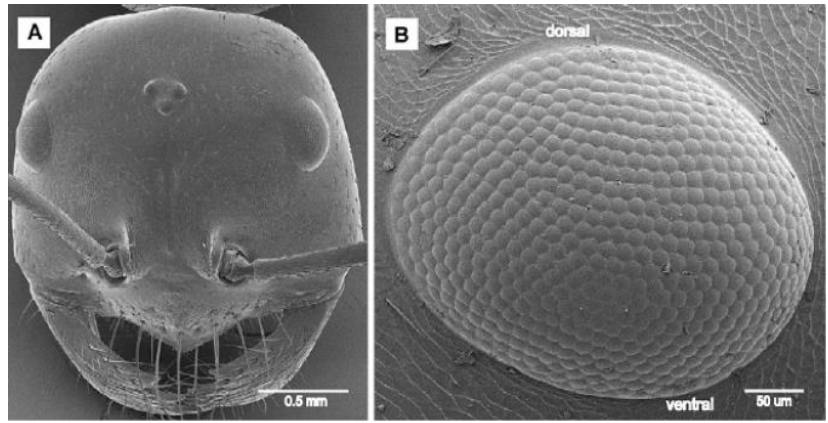

*Figure 1 (A) External morphology of the M. Bagoti head (dessert ant). (B) A scanning electron microscope close-up of the compound eye of the ant with between 421 and 590 ommatidia. (Schwarz, Narendra and Zeil, 2011)*

Another interesting fact about the navigation of the ants is that they use a peripheral visual system in order to perceive the visual information from their environment. Specifically, they possess typical apposition eyes with about 420-590 ommatidia per eye, a horizontal visual field of approximately 150º and facet/lens diameters between 8 and 19mm, depending on body size, with frontal facets being largest (Schwarz, Narendra and Zeil, 2011). In the majority of the insects including ants, the visual pathway begins with a compound eye. Compound eye receive information from multiple lenses with differing viewing directions (Land and Nilsson, 2012).

The structure of the visual system of the ants described above allows for a very large field of view but a low-resolution view. According to past researches wide field vision and mid-range visual resolution are better suited for navigation (Graham and Philippides, 2017).

The visual navigational behaviour of the ants is the proof that low resolution and small brain is not a limit for high performance navigation. This is also the reason why many researches focuses on the emulation of insect-level behaviours.

## 3.2 Snapshot Model and Chaining techniques

The navigational behaviour of many insects, including the ants, can be described using the "Snapshot Model". As it was mentioned above, ants find their way to a goal location by matching retinotopic views as remembered from their first traversal of the route.

Such models of visual navigation that have been successfully used for traversing foraging routes, are dominated by the snapshot-type model (Baddeley et al., 2012). According to the snapshot model, a panoramic image of the target position is acquired and stored by the animals. Subsequently, the stored snapshot is recalled and used to calculate the direction that the animal has to follow in order to face its goal position. This is done by applying a matching mechanism between the current retinal image and the stored snapshot (Lambrinos et al., 1998). Consequently, the route navigation problem is reframed as a search for familiar views in order to recover the heading towards to the goal location.

Many experiments as the one conducted by Cartwright and Collett with bees have confirmed the adoption of snapshot models in insects and the use of stored goal view images to recover the route heading (Cartwright and Collett, 1983). This has inspired many biologists and computer scientists who focus their field of research in developing homing models based on the snapshot model.

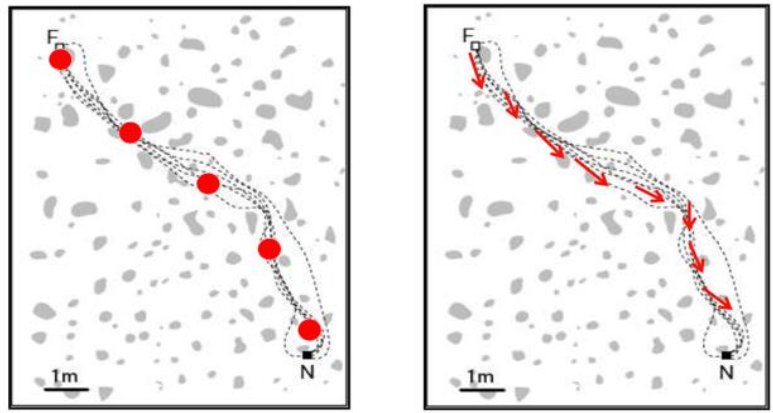

*Figure 2 This figure represents a chaining technique of the Snapshot model. As you can see in the left part of the image many snapshots were taken along the training route. At the right side of the image you can see that this model manages to recover the headings all along the route and only from a specific location (Philippides 2013).*

Although Snapshot model provides a robust performance, it presents a basic limitation. It is unable to handle more than a single view. This makes snapshot approaches suitable only for navigation in the immediate vicinity of the goal location, as they cannot achieve robust performance over longer distances (Baddeley et al., 2012). This also means that if an ant is forcibly moved away from its goal's location it would be unable to return back to its base as it cannot handle navigating through the whole route but only through a specific snapshot that it memorized.

This problem can be overcome by collecting a series of snapshots along the route linked together as a sequence (figure 2). This approach can be characterised as Chaining technique. According to this, the navigational behaviour takes place from one stored view to the next in a fixed sequence along the route (Philippides et al., 2015). Although this model of navigation can work, previous attempts have shown that the agent should be able to robustly determine at which point of the route should every snapshot be taken and identify whether or not it has reached a snapshot during navigation phase. This means that the agent should have a "place recognition system" in order to be able to identify where along the route it is located.

Despite their limitations, snapshot models have managed to replicate the navigational behaviour of the ants with high robustness and this is the reason that they have been used as baselines methods for more advanced approaches to the same problem.

## 3.3 Artificial Neural Networks approach

As it was mentioned in the previous section, Snapshot model is a successful model and manages to replicate the visual navigation behaviour of the ants accurately. In this model, the route is defined in terms of discrete waypoints (snapshots) that the agent should always have access to and use them as references in order to recover its heading and stay in route. However, remembering every single snapshot experienced across a training route is an unrealistic assumption and needs excess computational power. For this reason, Dr. Baddeley, Dr. Graham, Dr. Husbands and Dr. Philippides proposed a different model that refines and develops the basic ideas of the snapshot model in a more realistic and biological way. In this case, they used artificial neural networks to store all the views experienced during training providing a holistic representation of the route (Philippides et al., 2015). Specifically, they used three types of neural network to learn the regularities of the training views collected during a single route traversal, in a way that the networks output the familiarity of their current views. This approach reframed the problem of navigation in terms of a familiarity search of views that look similar with the training views (Philippides et al., 2015).

The basic idea of their approach is that they use the views experienced in the first traversal of the route as training data for an ANN, train it and produce a model that it is able to take as input the current view of the agent and output a familiarity score.

Their approach has a main advantage. Instead of learning distinct snapshots along the route, it uses all the views experienced in the training phase and use them to determine a measure of familiarity. In this way, they provide an effective way of storing the training data of the route that is also open-ended as it allows a possible incorporation of new information. Consequently, the agent does not need to select which views to learn (Baddeley et al., 2012).

### 3.3.1 ANNs as classifier

Their first attempt was to use ANN as classifiers. In order train a type of ANN like this, they had to generate positive and training data. For this reason, they collected views that were part of the route as well as views that are not part of the route. The positive views were those ones that forward-faced the route along the training traversal and the negative were those ones which faced the right and left direction of the movement at an angle of 45 º relative to the heading of the route.

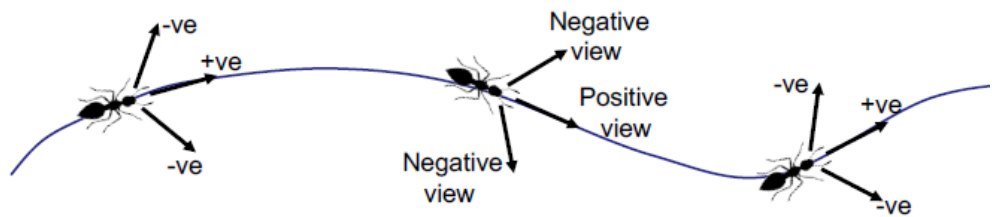

*Figure 3 Training Phase: The agent traverses the route and collects positive ("on-route") and negative ("off-route) training images.*

Before starting the training procedure, it was necessary to reduce the dimensionality of the input. The collected views had high dimensionality and it was a really difficult task to proceed with the classification. In order to lower the dimensionality of the data, they passed the views through a set of visual filters.

After reducing the dimensionality of the training data, they proceeded to the training phase. An efficient technique used in this case was Boosting which is a supervised learning technique for constructing a strong classifier by combining a set of weak classifiers trained with the labelled positive and negative examples. Specifically, they used Adaboost classifier. In this case, they initialized a big pool of 5000 different filters/classifiers of a random type, size and at random position of the visual field and they used the Adaboost classifier in order to select the most useful features. In order to control the complexity of the classifier they also set how many features the classifier can extracted by specifying their number (Philippides et al., 2015).

The results of this implementation proved that the more classifiers they used, the more accurate the performance of the ANN was.

### 3.3.2 Training without negative examples

Another one more biological and realistic attempt was training an ANN without negative examples. In this case, they collected just the views that they were part of the route. So, there were no negative examples to indicate if the agent is "off" the route. Instead of this, there was a powerful implementation which could accurately predict if a current view was "on" the route by learning a compact approximation of the distribution of the current views collected at the training phase.

In this case, they used Restricted Boltzmann Machines (RBM) an efficient implementation of "autoencoder" networks. After the RBM has been trained, it can use the current view as input and output its probability of being a part of the training set. The problem of navigation in this case has been reframed in terms of scanning the environment and searching for the viewing direction with the highest probability (Philippides et al., 2015) .

### 3.3.3 Learning and discarding views

The final and more advanced implementation is the Learning and discarding views. This approach is the current method used by the University of Sussex research team. It is similar to RBM approach as it does not use negative training views but requires much less training time. In this case, the collected views are used to train a two layered ANN to perform familiarity discrimination using Informax learning rule.

The difference between this and the previous ANNs techniques is that in "Learning and discarding views" approach, each training view is presented to the network just once and then discarded following a single cycle of weight updating. Therefore, the memory needed to store the training views remains constant.

This approach has a robust navigational behaviour and is also able to learn multiple routes to a specific goal. The performance of this method can be improved by increasing the number of the training route traversals (Philippides et al., 2015).

# 4. Methods / Results

In this section, we describe the environment and the basic methods of the project. After explaining the key features of the database/virtual environment that was used for this project, we analyze the Visual Compass and the Perfect Memory approaches which are inspired from the Snapshot model. These two methods were the baseline methods for our ANN approach. Finally, we describe our approach of using ANNs as classifiers and examine some specific aspects of its performance.

## 4.1 Virtual Environment / Database

As it was already mentioned, this project attempts to simulate the navigational behavior that ants exhibit, in an artificial world by implementing different kinds of biomimetic algorithms. In order to implement and test our models we used an image database of a simulated world provided by the University of Sussex. This database contains 2601 images from different positions on the grid of the simulation world (world grid images) as well as some training routes. The training routes include a sequence of images that they were taken from different positions from the positions of the grid's images and they are numbered according to their position along the route.

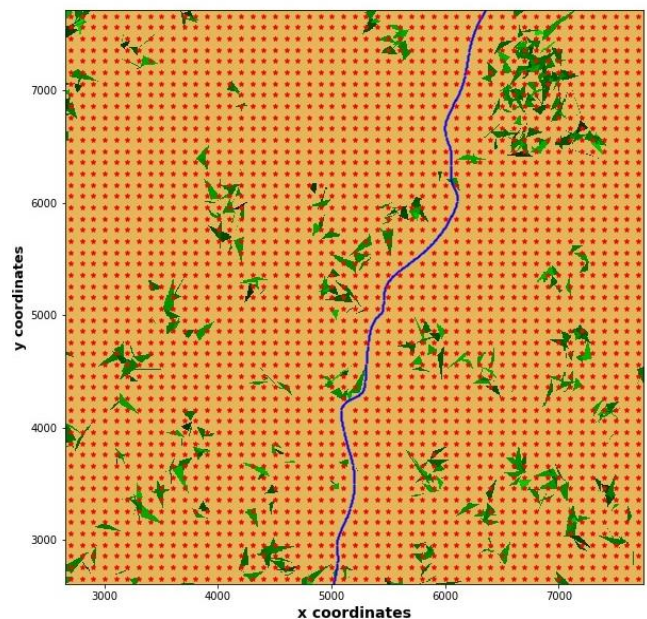

*Figure 4 Top view of the simulated world. The red dots represent the different world grid images and the blue line an example training route.*

The simulated world has been created to resemble the natural environment of the ants as it contains objects that would appear in an ant's environment like vegetation, tussocks and leaves of various sizes and shapes. According to previous researches, ants use panoramic views for route navigation (Schwarz, Narendra and Zeil, 2011). For this reason, the database contains panoramic unwrapped images like the one pictured in figure 5.

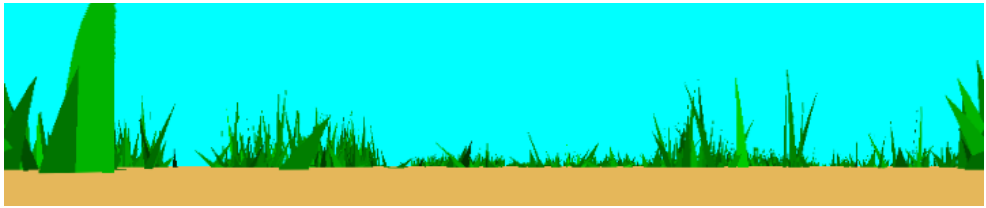

*Figure 5 Example of an image in the simulated world database.*

Finally, the images of the database are not only determined by their location but also by the heading of the agent when it took the picture. The heading parameter is set in degrees and expresses the angle that an image should have in order to face the goal location.

## 4.2 Visual Compass

The first and simplest model that was implemented is the Visual Compass. This model is based on the Snapshot Model that was analyzed in the background section. The basic concept of Visual Compass is that the heading of the agent when a training view was taken can be precisely recovered by comparing the current view with the training image.

Ants face the route direction most of the time during a learning walk. The views experienced along their learning walks define the movement directions required to stay on the route. So the problem of navigation is reframed in a rotational search of familiar views. This is what exactly our model, Visual Compass, tries to replicate.

### 4.2.1 Image Difference Function

In order to compare and identify the differences between the route image and the image from the world grid we built an image different function IDF. This function takes as arguments these two images and depicts how the current image changes with difference from the goal view.

The IDF's formula is given below:

$$IDF(current\ image, goal\ image) = \sqrt{\sum_{i=0}^{P}(C_i - G_i)^2 \Big/ P} \qquad (1)$$

Where P value is the number of pixel of the images, $C_i$ is the $i^{th}$ pixel of the current image and $G_i$ is $i^{th}$ pixel of the goal image. The output of the IDF is the root-mean-square (RMS) value of the difference between the two images. If the images are exactly identical this value will be zero. This means that the more similar the images are, the lower the RMS value will be.

In order to find the best matches, we should rotate the current image through 360º and iteratively call the IDF function to calculate the root mean squared error between the goal and the rotated current image. The best match is the rotated image with the lowest root mean squared error.

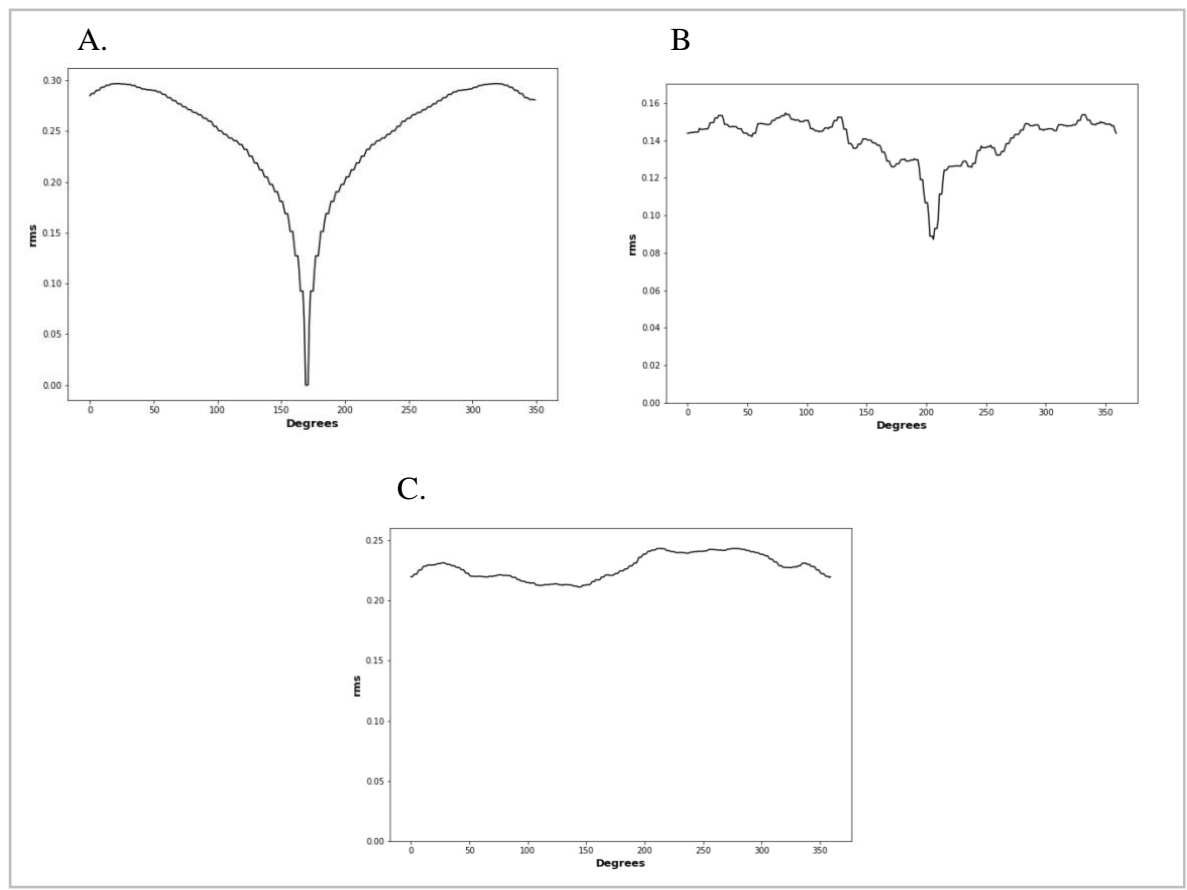

*Figure 6 This figure depicts three different IDF Diagrams (rms vs degrees). The diagram A is an IDF diagram when the current image = goal image. The diagram B is an IDF diagram when the current image is close to the goal image and the diagram C is the IDF diagram between a goal and a current image where the current image is far away from the goal image.*

The previous figure depicts three diagrams produced by the Image Different Function for three different cases. All of the diagrams plot the root mean square error between a goal and a current image regarding to the rotated degrees of the current image. In this example we rotated the current image by 360º.

The first case is when the current image is equal to the goal image. In this case, the diagram shapes a "V" finding a peak minimum of zero RMS error. This case is represented by diagram A. The second case is when the current image is similar to the goal image. In this case the IDF diagram will also shape a "V" but the minimum will never be zero as the current image has differences with the goal image. This case is represented by diagram B. Finally, the last case is when the goal image is entirely different from the current image. In this case the diagram will form irregular curves and it will not have a distinct peak minimum as in the previous cases. This case in represented by diagram C.

The use of the Image Different Function in our model is to find where the root mean squared error between the current and the goal image becomes minimum and identify how many degrees the agent should rotate in order to face the goal.

### 4.2.2 Visual Compass results

As it was already mentioned the database we use has two different types of data. Images from different positions of the virtual world and example route images. The example route images, which are also the training data, have a set heading in order to face the goal view. Using the IDF function we can calculate how many degrees the current image should be rotated in order to face the goal view as well. If the goal image and the current image have been taken from similar locations, the IDF should calculate that the heading for the current image should be similar to the set heading of the training image.

A good representation for the heading is a vector diagram as the diagram shown in figure 7. These diagrams can be used to represent the direction to the goal view from a specific location on the grid.

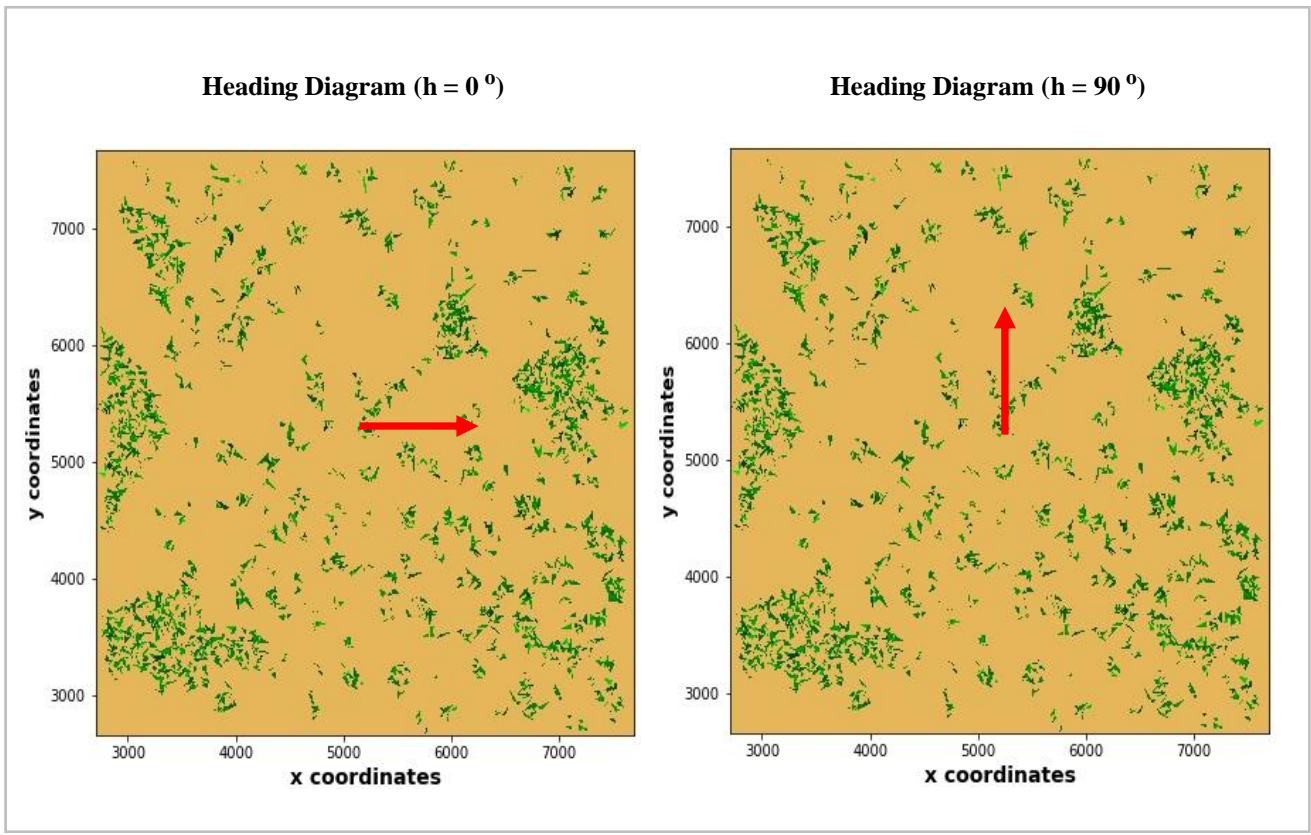

*Figure 7 These figures represent two different headings for the same location (x, y) = (5301, 5159). The left diagram represents a heading of 0° and the right heading represents a heading of 90°.*

This figure represents two different headings for the same location of the world grid. Specifically, the left diagram represents a heading of 0° and the right diagram represents a heading of 90°.

The basic idea of Visual Compass is that it uses a reference training image, with a set heading to face the goal, to recover the heading for the current image by using the procedure described above.

The next figure represents a multiple heading diagram between a specific reference image and images of the world grid. We selected to plot the heading diagram for world grid images in a distance range of 500mm around the goal image. The reason for setting up this limit is to focus on the points of the grid which are close to the image of the route.

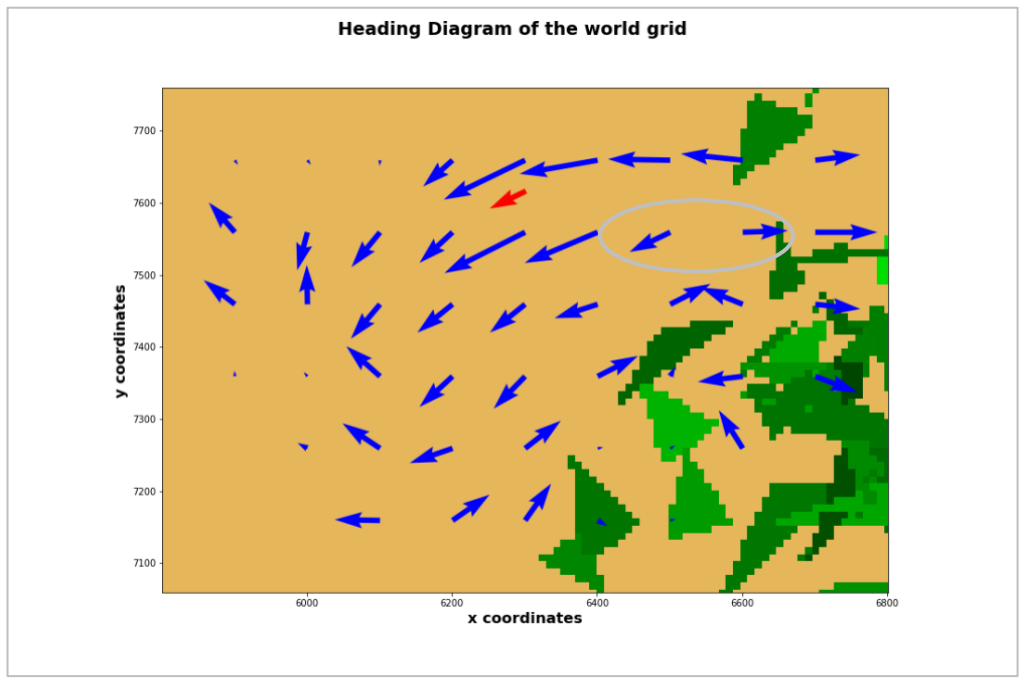

*Figure 8 This is a multiple heading diagram. The red arrow depicts the heading of the route image with coordinates (x, y) = (6215.72, 7371.97) and the blue arrows depict the recovered heading of the different images of the world grid. The length of the blue arrows is inversely proportional to the minimum rms value. That means that the lengthier the arrow is the more similar the world grid image and the goal image are.*

The graph contains the heading of the goal image coloured in red and the recovered headings of the images of the grid coloured in blue. Also, the length of the arrows which represents the recovered headings, has been set to be inversely proportional to the minimum rms. That means, that the smaller the minimum rms error is, the lengthier the arrow will be. This representation helps to immediately understand the similarity between the goal image and the rotated image of the world grid.

According to the graph, our model manages to successfully recover the headings for the majority of the grid images which are located near to the goal image. However, there are some cases that visual compass may produce some unexpected results. For example, in the right up corner there are two matches from nearby positions that point in different directions (the two blue arrows which are circled in figure 8. Specifically, the arrow at the left managed to recover the correct heading while the arrow at the right did not recover a similar heading with the heading of the reference image.

In order to examine what happens in cases like this, we apply the visual compass model for these two world grid images separately. In figure 9, we can see the Image Difference Function

Diagram for the both of the world grid images. According to this diagram, it is clear that the IDF function diagram for the left world grid image has one peak minimum. On the other hand, the IDF diagram for the right world grid image found more than one minimum and did not recover the correct heading for the specific world grid image

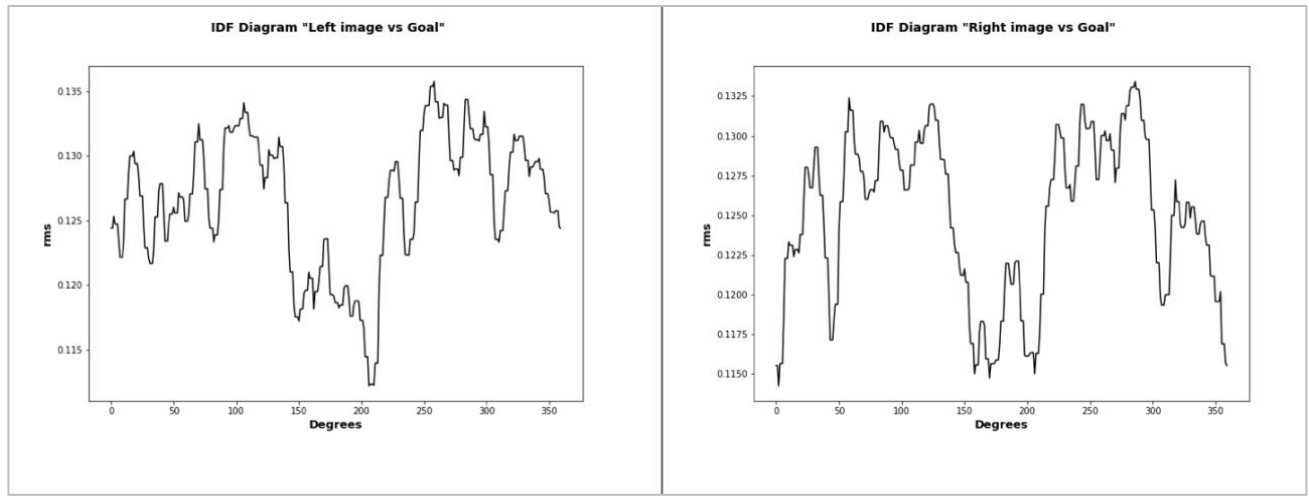

*Figure 9 This figure represents the IDF diagram between the two world grid images and the reference image. The left diagram is the IDF of the left image and form a distinct peak minimum. The right diagram is the IDF of the right world grid image and forms more than one minimums.*

The reason for this is that the environment of the ant may look similar from different rotations. In this case, as we can see from the diagram, the IDF found 4 different minimums including the correct one. However, the environment from this location looked very similar from all these different directions and the IDF did not manage to recover the correct heading.

Another common case in which the visual compass produces inaccurate results is when a possible obstacle (tussock, leaf) is inserted between the world grid image and the goal location (figure 10).

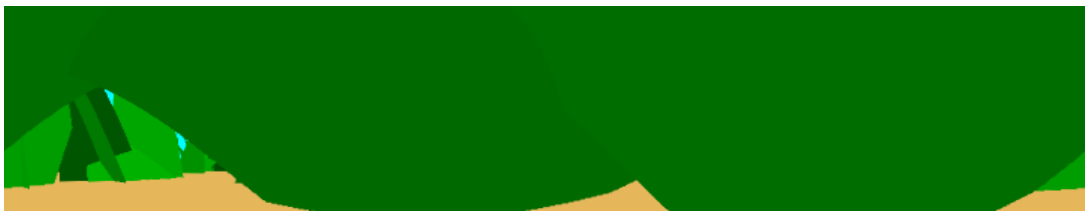

*Figure 10 This figure represents an example in which a tussock of the environment affects to the field of view of the agent.*

In cases like this, the inserted obstacle produces noise to the agent's field of view and as a result the IDF function cannot manage to recover a correct heading.

Finally, there is another possible problem which is related with the the distance between the current and the goal view. As we can see in the figure 8 the recovered headings of the world grid images which are located away from the goal view are not similar with the reference heading. Also the length of these arrows is really small indicating that their rms error is high. That means that these images have many differences from the goal image and as a result the model cannot recover the correct heading for them. These data can be considered as outliers.

Although visual compass may produce some inaccurate results, for the majority of the images which are close to the reference route image, manages to recover a similar heading with the heading of the route. For this reason, we can conclude that this model has a robust performance and can be used as a baseline method for the artificial neural networks implementation.

### 4.2.3 Error in Visual Compass

Another way to identify how robust is the Visual Compass is to calculate its error. The error of the model is the absolute difference between the heading of the reference image and the recovered heading of the world grid image.

$$error = |hs - hr| \qquad (2)$$

Where hs is the set heading of the training image and hr is the recovered heading of the world grid image.

In general, the closer the world grid's images are to the reference image, the smaller the heading error will be. This is happening because in most of the cases, the goal image will have many similarities with the images around its location. Consequently, the IDF will be able to recover a heading really close to the set heading of the route image

This statement can also be confirmed by plotting a "heading error vs distance" diagram. In order to produce a diagram like this we should iteratively call the IDF function between a specific reference image and different images of the world grid. From the produced graph in figure 11, it is clear that there is a trend, as the heading error values increasing with the distance from the goal image.

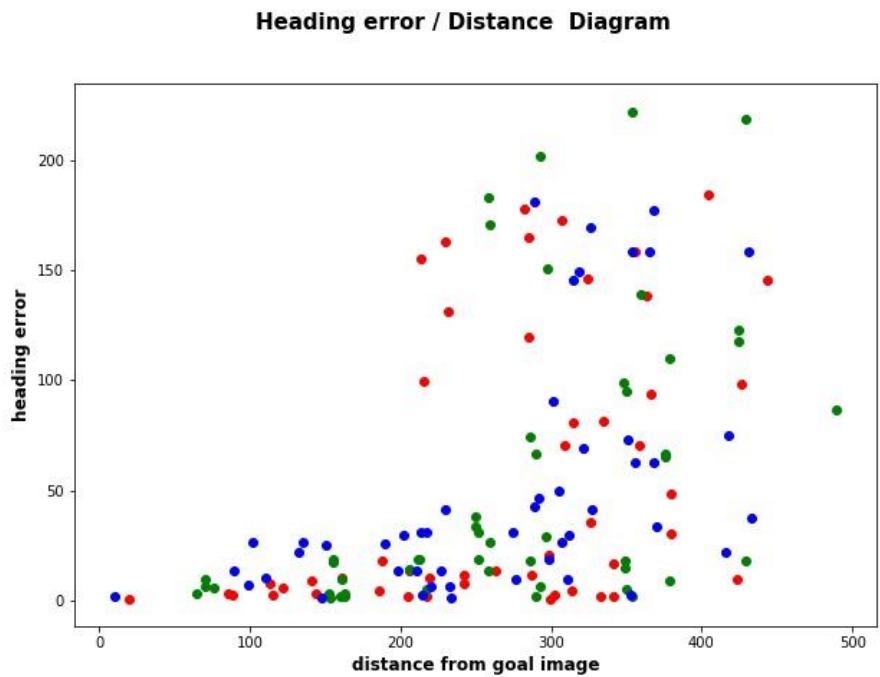

*Figure 11 This figure represents a heading error vs distance diagram. Each colour represents a different case of a reference image. In all of the cases the heading error values between the reference images and the images of the world grid increase with the distance from the goal image.*

Specifically, the diagram depicts three different cases, each one represented by a different color. In all of these cases, the heading values between the corresponding reference image and the images of the grid increase with the distance from the goal image. We selected to plot the heading error between the reference and the world grid images in a distance range which does not exceed the 500mm. The reason for setting up this range is because in a longer distance the world grid's images which are far away from the goal image produce noise in the graph and the relationship between the heading error and the distance from the goal does not become clear.

## 4.3 Perfect Memory

Although visual compass has a robust performance, it presents the basic limitations of the Snapshot model. Specifically, it cannot handle more than a single view and as a result, it cannot be used for navigating through the whole route. For this reason, we implemented a new model called Perfect Memory. This model is also based on the Snapshot Model but instead of handling a single snapshot, it tries to make a chain of all the views experienced along the training route. Perfect memory was used to investigate the preprocessing and the frequency of the training data that we used in our ANN implementation.

### 4.3.1 Perfect Memory Implementation

Perfect Memory is a simple and complete implementation imitating the visual navigation system of the ants. This approach relies on the unrealistic assumption that the agent can perfectly memorize all the views experienced along a training route and determine view familiarity using the Visual Compass model. Specifically, this solution follows these basic steps:

1. Store the training views captured along a training route

2. Apply Visual Compass between the current image and all the training images and store the minimum outputs of the IDF.

3. Recover the heading of the current image according to the smallest minimum of the previous process.

This model reframed the problem of navigation as a similarity search between the current view and the training views of the agent. In the figure below, you can see the recovered headings of the world grid images along a training route using Perfect Memory model. We selected to recover the heading for the world grid images, which are located in a range of 200mm from the route.

**Recovered Headings Perfect Memory**

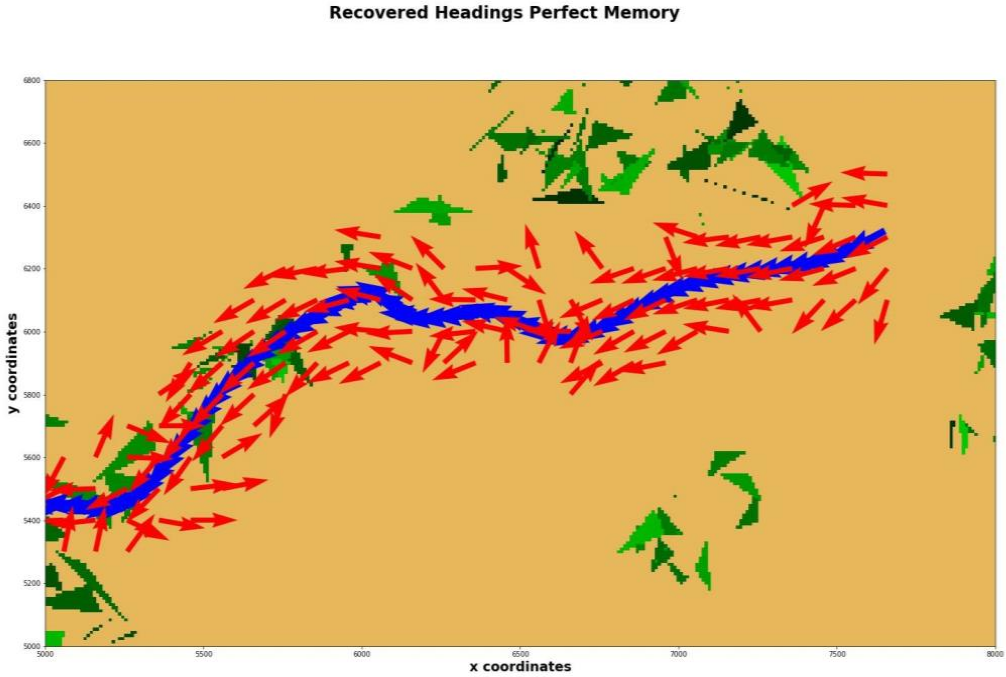

*Figure 12 This figure represents the recovered headings of the world grid images along the training route. The blue arrows represents the set heading of the route and the red arrows represents the recovered headings.*

The reason for setting up this limit is that as we proved in the error analysis section of the Visual Compass, it cannot recover the headings for the world grid images in long distances from the goal image. As we can see from the heading diagram (figure 12), Perfect Memory model manages to recover the headings for the majority of the world grid images. However, in some cases it fails to recover the headings accurately. This is happening because our model is based on Visual Compass and as a result it presents the same limitations.

### 4.3.2 Error in Perfect Memory

In order to evaluate the performance of the Perfect memory model we used a similar error calculation system as we did in Visual Compass. Specifically, the error of the model is the average absolute difference between the headings of the reference images and the recovered headings of the world grid images. The error calculation formula is given below:

$$Perfect\ Memory\ error \quad = \quad \sum_{i=0}^{n} \left( |hs_i - hr_i| \right) \Big/ n \qquad (3)$$

Where the hs is the set heading of the training images, hr is the recovered heading of the world grid images and the n is the number of the recovered headings.

One problem that we had to overcome was finding the corresponding set heading for every recovered heading. In order to overcome this problem we implemented a "place recognition system which finds the closest snapshot for every world grid image. Using this method, our model is able to calculate the heading error between every current and training image.

**Place Recognition System Representation**

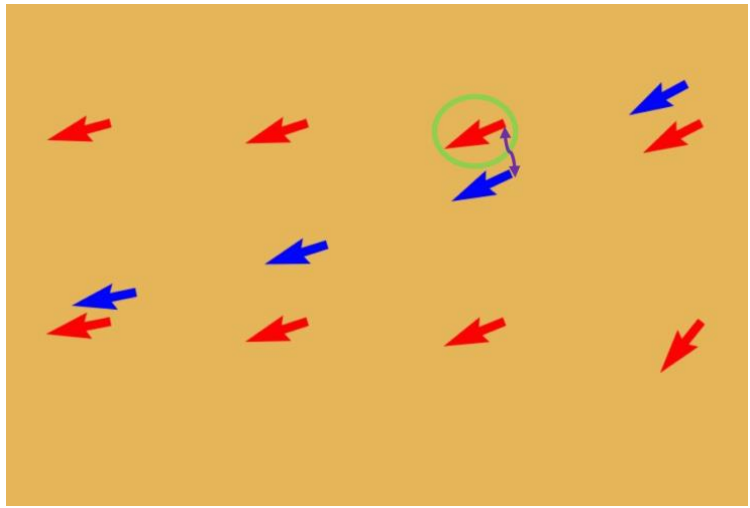

*Figure 13 This figure represents the way that our place recognition system works. As you can see in the picture, the circled recovered heading will be correlated with the closest training image.*

Another observation that we did using this error calculation system is that the average error underestimates the accuracy of the Perfect Memory model. That is because an average heading error can be significantly biased by the outlier data. In our case, these data can be considered the headings that Perfect Memory completely failed to recover. For this reason, a more representative error calculation is calculating the median absolute difference between the headings of the reference images and the recovered headings of the world grid images.

### 4.3.3 Using Perfect Memory to specify the Frequency of the training data

Perfect Memory will be used as a tool to investigate how the frequency and the preprocessing of the training data affect the performance of the model. The aim of this investigation is to specify the number and the resolution of the training data that should be used to train our Classifier.

In general, the more training data used the higher accuracy the neural network produces. However, we should try to reduce the number of training images fed in the neural network in order to be able to produce results in a reasonable amount of time. For this reason, we tested the perfect memory using different frequency of data, meaning that we set Perfect Memory to take into account 1 training image every n training images, where n stands for the frequency value. Specifically, we set different frequency values and we calculated the corresponding heading errors for the whole route.

Following this approach, we produced the next diagrams, which depict the relationship between the frequency of the training data and the error of the model. The first diagram (figure 14) represents the median error of the Perfect Memory model for different frequency values and the second diagram is a whisker box diagram (figure 15) which expresses the same relationship between the overall error of the model and the frequency of the training data.

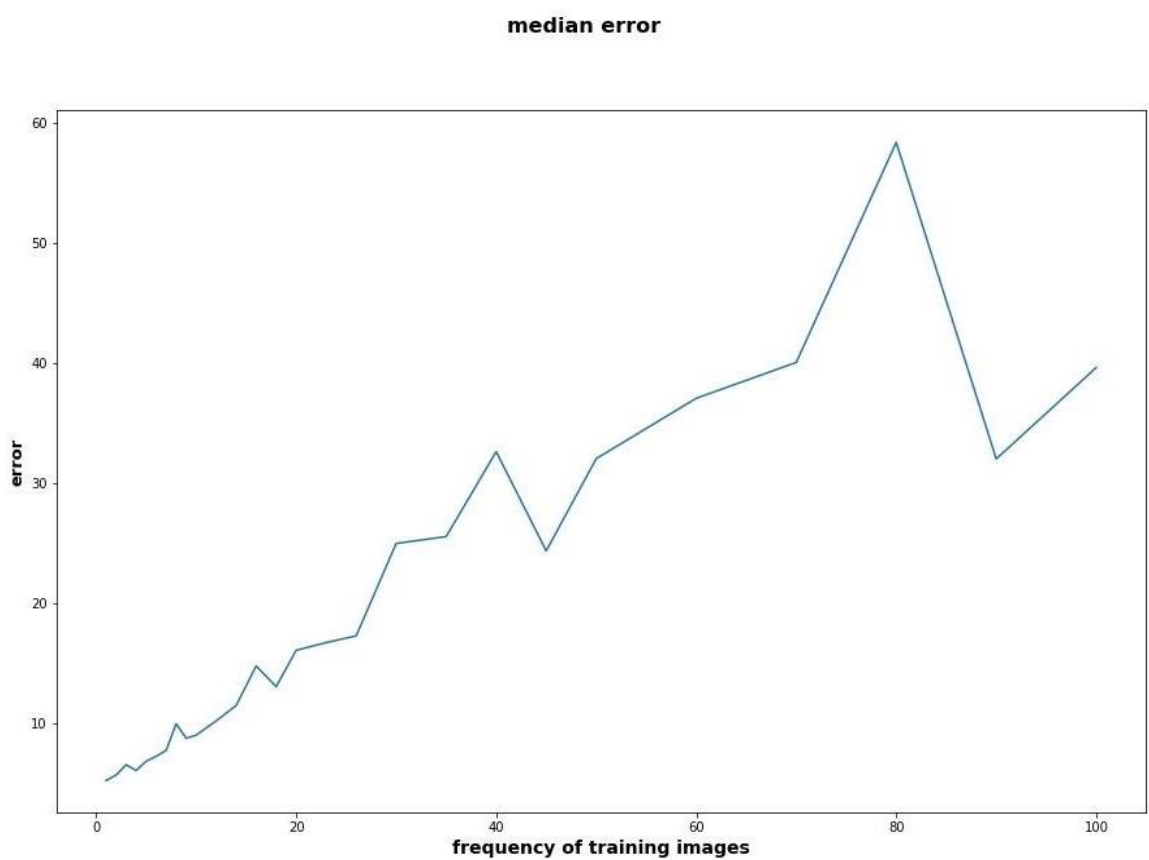

*Figure 14 This figure represents the relationship between the error of the Perfect Memory model and the frequency of the data. As you can see, the less training data used the more the median error of the model is.*

**Whisker box Diagram**

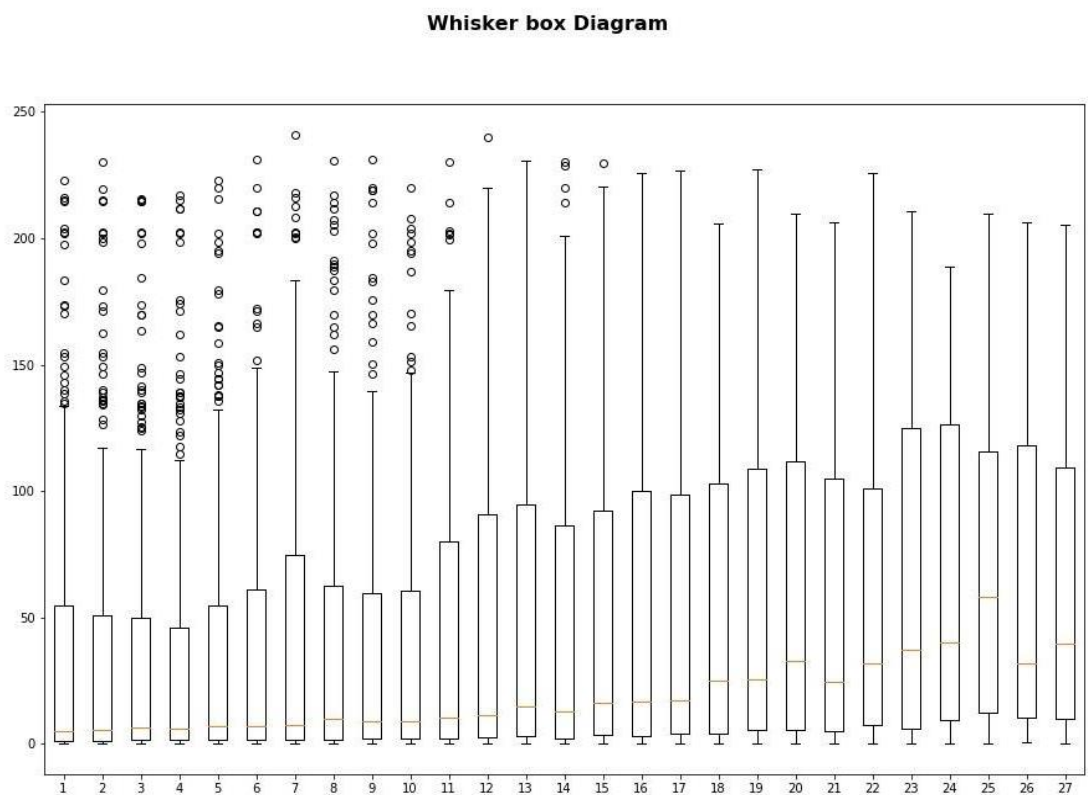

*Figure 15 This figure is the whisker box diagram between the error and the frequency of the training data. This representation is really informative as it provides information about the minimum, maximum, standard deviation and the median error of the model as well as the outlier data.*

According to the produced error / frequency of training data diagram, the error of the model has an inversely proportional relationship with the number of the training data. That means that the more training data used the less error the model produces. Looking at the Whisker Box diagram we see that the median error increases linearly until the $20^{th}$ frequency value. In order to select the most optimal and reasonable value we checked out the distinct error values of the model (Appendix 7.2) as well as the error distribution among the whisker's box quartiles. After analysing the results, we concluded that a frequency value of 6 is a good compromise as it produces a relatively small median error value and low standard deviation.

This decision means that we will test our ANN implementation using one route snapshot every six route snapshots. This will help us to train our neural network in a reasonable amount of time and produce relatively accurate results.

### 4.3.4 Using Perfect Memory to specify the Preprocessing of the training data

The second aspect that we will analyze with Perfect Memory is how the resolution and the preprocessing of the training data impact on the overall error of the model. The reason for this investigation was to identify what preprocessing techniques we should use in our ANN implementation in order to reduce the dimensionality of our training data. The preprocessing techniques that we will focus on are:

*Converting RGB image into greyscale:*
RGB images contains lots of data which are not required for our processing. An easy and clever way to get rid of these data is to convert RGB image into Gray scale. This was also the first preprocessing technique that we applied to our data.

*Image Resizing:*
Image resizing is a classic preprocessing technique. The effect of this technique is blurring. Using this technique to preprocess our data will help to reduce the image dimensionality and convert the images into low resolution. Reducing the resolution of the images will also help to produce a precise biological model of ant's visual navigational system as according to the background section, they also perceive the world in low resolution.

*Gaussian smoothing*
Gaussian smoothing is widely used method to reduce the image noise. Specifically, we will use it as a preprocessing technique in order to enhance the structure of the image data.

In order to find out the best combination of the preprocessing techniques described above, we will use Perfect Memory model. Specifically, we will calculate the overall median error of the model for different resizing factor values.

As shown in the median error vs resizing factor diagram in figure 16, minimum error occurs for a resizing factor of 6. Before this point the error is relatively stable while for higher resizing factor values the error starts to increase exponentially. For this reason, we chose to resize our images by a factor of 6.

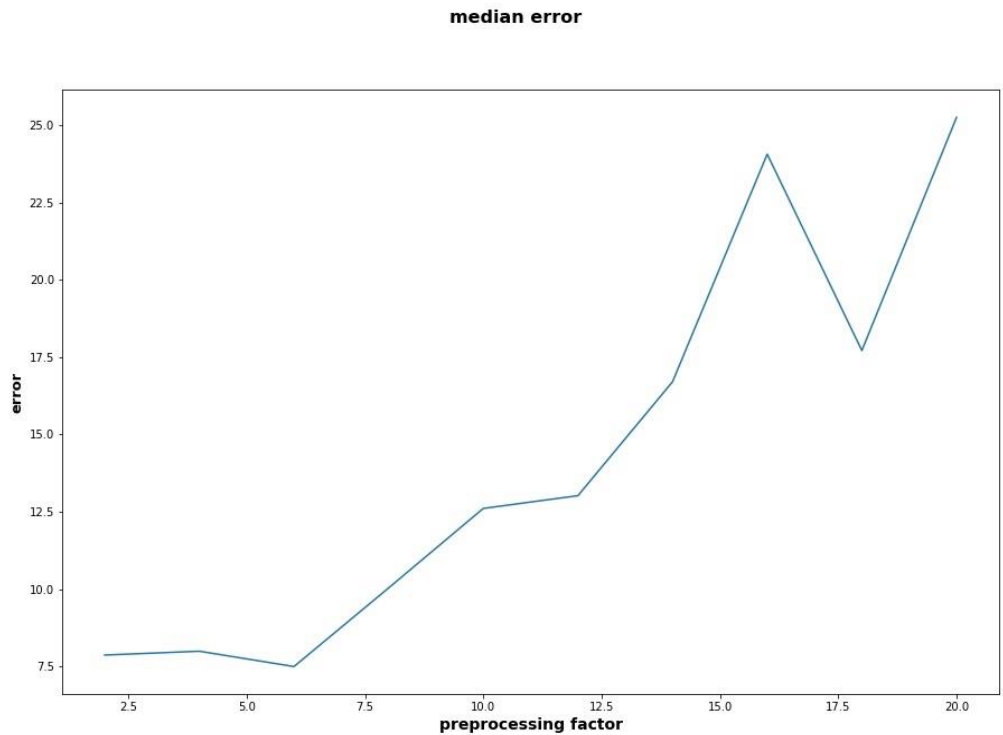

*Figure 16 This figure represents the relationship between the median error of the model and the resizing factor of them model.*

According to this diagram, using a mid-range resolution produces better results than using the original resolution of the images. This outcome confirms the theory that wide field vision and mid-range visual resolution are better suited for visual navigation which was also stated in the background section. In order to produce the diagram in figure 16 we made use of Gaussian smoothing as well. After calculating the error of the model with and without this pre-processing technique we concluded that it finally improves the performance of the model and decreases its overall error. You can see the rest of the produced graphs and their analysis in the Appendix 7.3.

Perfect Memory was the baseline method of our project as well as a very useful tool to investigate how the resolution and the frequency of the training data impact on the performance of the model. This investigation helped us to specify the optimal frequency and pre-processing that we should set in our training data in order to produce relatively accurate results in a reasonable amount of time using our "ANNs as classifiers" implementation. To sum up, the analysis of the results concluded that the most optimal tuning for our data in order to use them in the ANN approach is: frequency value of training data = 6, resizing by a factor of 6 and Gaussian smoothing.

## 4.4 ANNs as classifiers

Perfect Memory is the first complete biomimetic algorithm that we implemented. Although, it produces high accuracy results it is based on the unrealistic assumption of remembering every single snapshot of the route. At this section we will approach the same problem with a more advanced approach, Artificial Neural Networks as Classifiers. Specifically, we will use a Multi-Layer Perceptron (MLP) classifier. Using the results of the frequency and preprocessing investigation that we produced with the Perfect Memory model, we will now try to explore how the MLP classifier approaches the same problem. The research will focus on how the number of the training data impact on the accuracy of the model and why our model performs better in specific parts of the route than others.

### 4.4.1 Data Selection / Preprocessing

As it was already described in the previous section, Perfect Memory was the baseline method for finding the optimal values and specify the frequency and the preprocessing of the training data. According to this, the frequency of data that it should be used is 1/6 (one snapshot every six). This selection will help our ANN implementation to produce relatively accurate results in a reasonable amount of time.

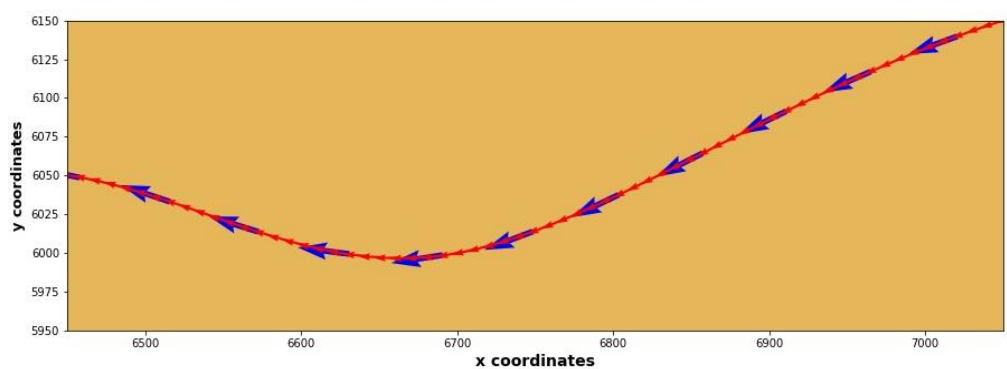

*Figure 17 This figure represents the frequency selection of the training data. The blue arrows represent the selected training data and the red arrows represent the discarded snapshots.*

After selecting the correct frequency of the training data we proceed in the preprocessing phase. According to the results of the Perfect Memory investigation, the optimal preprocessing for our training data is converting the images to gray scale to reduce the dimensionality of the data, Gaussian smoothing to reduce the noise of the image data and Image resizing to reduce the data dimensionality and produce low resolution images. The results of the Preprocessing techniques applied in our data depicted in the next figure.

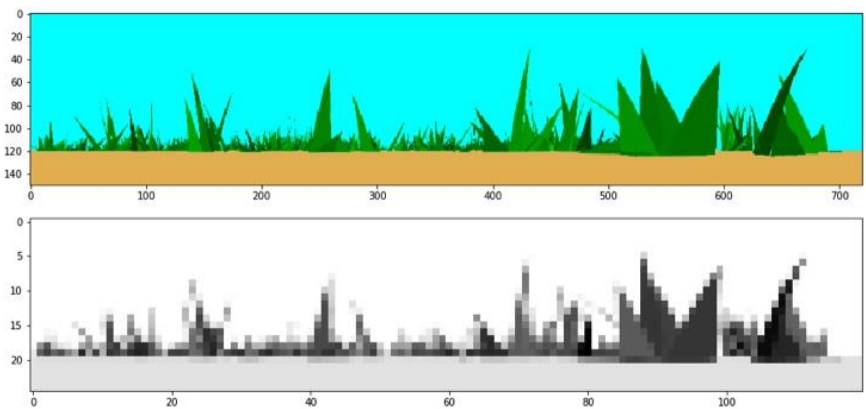

*Figure 18 This figure represents an example training image before and after the pre-processing phase.*

### 4.4.2 ANNs as classifiers implementation

The basic idea of the "ANNs as classifiers" approach is that the route snapshots are used as training data to train the Classifier and use the trained model to produce a familiarity score for every word grid image indicating whether is facing the route or not. Our approach can be divided into two separate parts, the training phase and the action phase.

*Training Phase:*

In order to train the classifier, they were used positive and negative training data of the route. Positive are the images that face the route and negative are the images that do not face the route. In order to produce our training data, we use the route snapshots as the positive training data and the rotated versions of those snapshots for the negative ones.

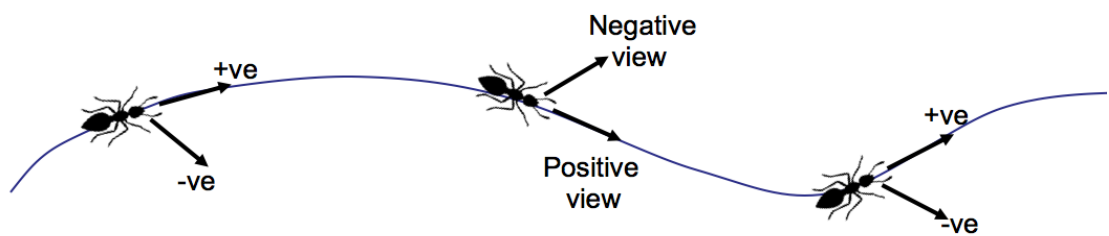

*Figure 19 This figure represents the way that we collected the training data. As you can see in the picture the positive training data are those ones which face the training route and the negative are those ones which face the right and left direction of the movement at an angle of 60º relative to the heading of the route.*

Specifically, as it is depicted in figure 19, for every positive training image we generate the corresponding negative training image by rotating the snapshot 60 º right and left from the direction of the route alternately. The reason for this is that we want to produce same number of positive and negative training data.

After collecting our positive and negative images, we use them as labelled training data to train our classifier. We decided to set the labels 1 for the "on route" training images and 0 for the "off route".

_Action Phase:_

After the training phase our model is able to take as input the current view of the agent and output a familiarity score indicating whether this view is part of the route or not.

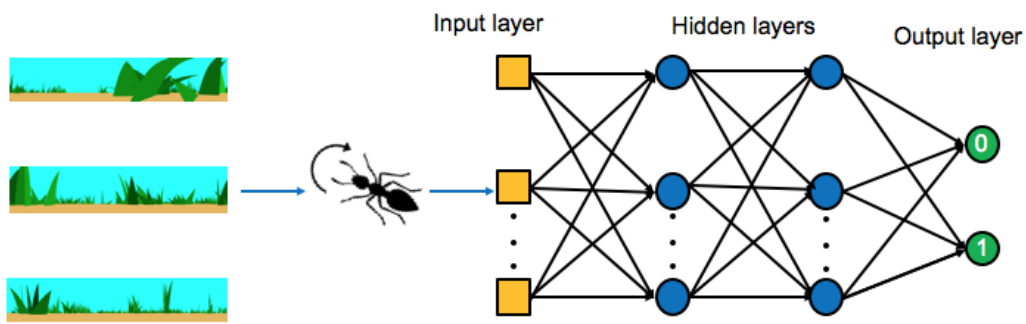

_Figure 20 This figure represents the action phase of the ANNs as classifiers. Specifically, the agent scans the environment by rotating around itself, inputs the views it sees to the ANN and moves to the direction with the highest classification familiarity "on route" score._

In order to identify the correct direction, the agent follows a similar procedure with the one described in the Visual Compass model. This procedure follows the next steps:

1. The agent scans the environment by rotating around itself 360º, takes its current view and uses it as input in the ANN.
2. The ANN takes as input the current view of the agent and output a familiarity score indicating whether this view is part of the route or not.
3. The agent recovers its heading according to the highest familiarity score of the ANN.

The next diagram represents the familiarity score produced by our ANN for the world grid image with coordinates (x, y) = (6959, 6101) through a 360º rotation. In order to produce this diagram, we used our place recognition system, described in Perfect Memory section to identify the closest route snapshot. Using this snapshot, we produced a positive and a negative view to use them as labelled training data for the MLP. Finally, we used our trained model to produce the familiarity score values for a 360º rotation of the current image.

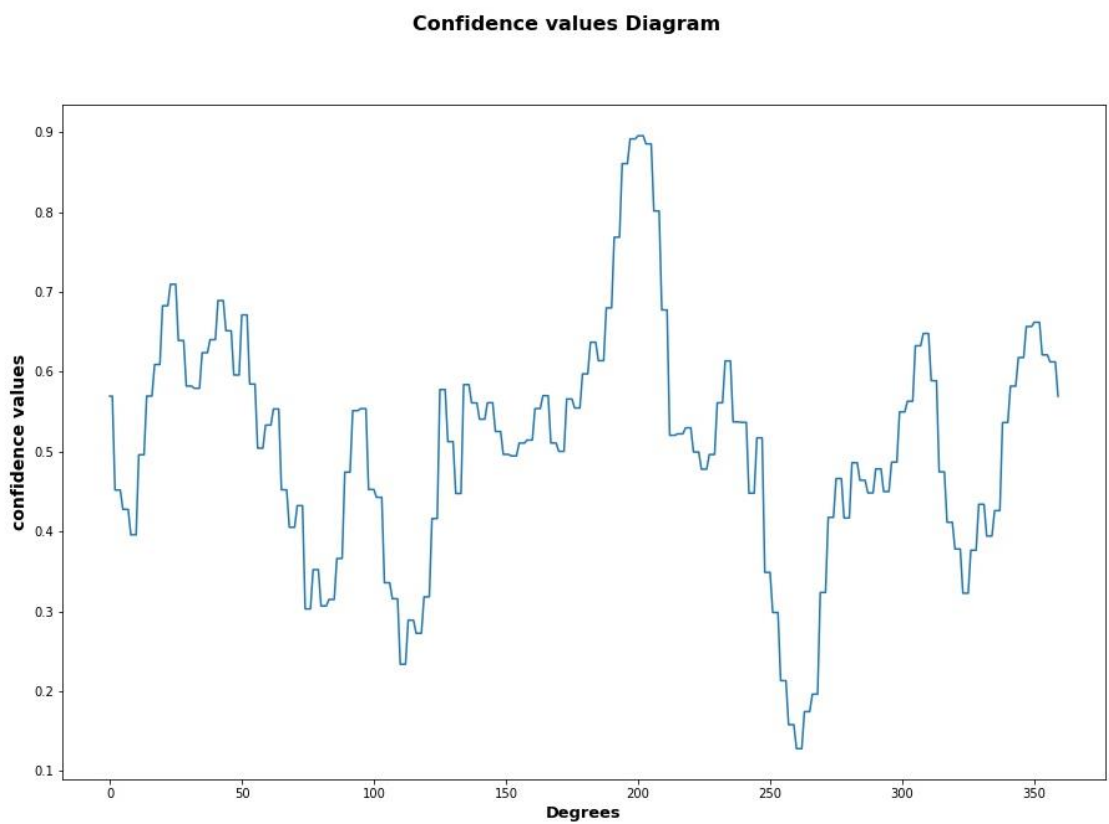

*Figure 21 This figure represents the familiarity score produced by our ANN through a 360º rotation of the world grid image with coordinates (x, y) = (6959, 6101). According to the diagram the familiarity score becomes maximum for a rotation of 203º.*

As shown in the Confidence values vs Degrees diagram, the familiarity score becomes maximum for a rotation of 203º. That means that the current image should be rotated by 203º to face the route. In order to confirm this result, we also produced the Heading diagram between the training snapshot's heading and the recovered heading of the world grid image.

According to the produced Heading diagram in figure 22, our model managed to recover the heading successfully. Specifically, the error between the set and the recovered heading was only 0.35º.

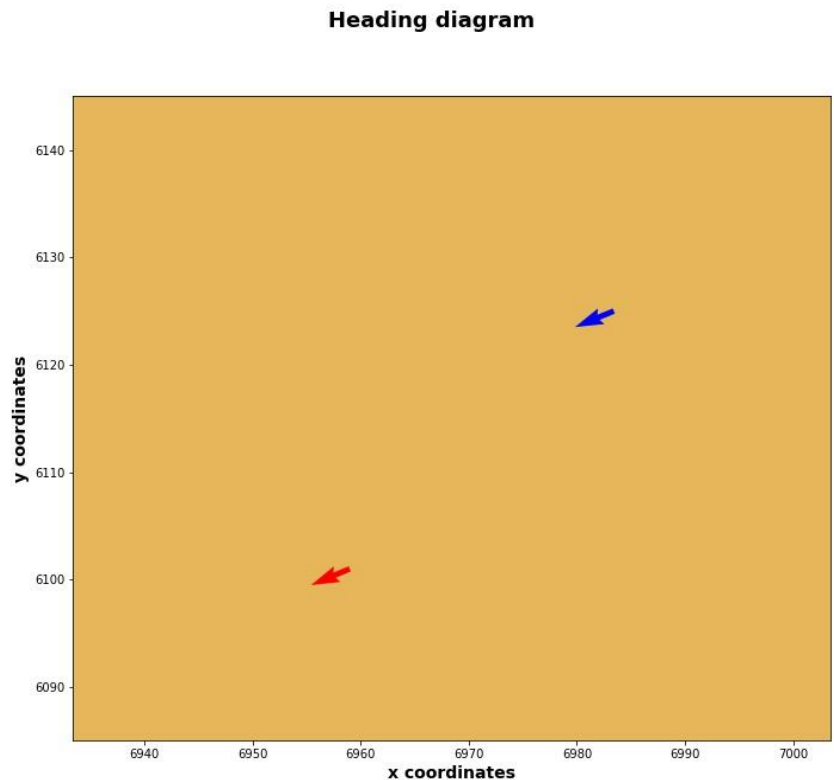

*Figure 22 This figure represents the heading diagram between the set heading of the snapshot of the training route and the recovered heading of the world grid image with coordinates (x, y) = (6959, 6101). The blue arrow represents the heading of the training image and the red arrow represents the recovered heading of the world grid image.*

The procedure described above was the basic idea of our ANNs as classifiers implementation. However, in this case we used only one snapshot to train our model and recover the heading for the closest world grid image. In order to do this task, it was not necessary a more sophisticated tuning of the ANN or any extra investigation as we had a really small dataset. This example was presented to explain the procedure of the ANNs as classifiers approach.

In the next section of the report we will investigate how this approach performs for a bigger number of training data, what tuning we should use to our classifier as well as in which parts of the route the ANN cannot perform well and why.

### 4.4.3 Evaluation of our ANNs performance

Before testing the MLP classifier, it was required to specify how to evaluate its performance. We decided make a double dimensional evaluation by testing both the accuracy of the classifier as well as the median error between the recovered and the set headings.

_Accuracy evaluation:_

In order to calculate the accuracy of our model it was necessary to collect some extra labelled data and use them as a test dataset. Test dataset is a dataset that is independent of the training dataset and is used to provide unbiased evaluation of the trained model. In this case, we had to collect some "on route" and "off route" views that the model has never seen before. We decided to collect these data using the intermediate snapshots of the training data.

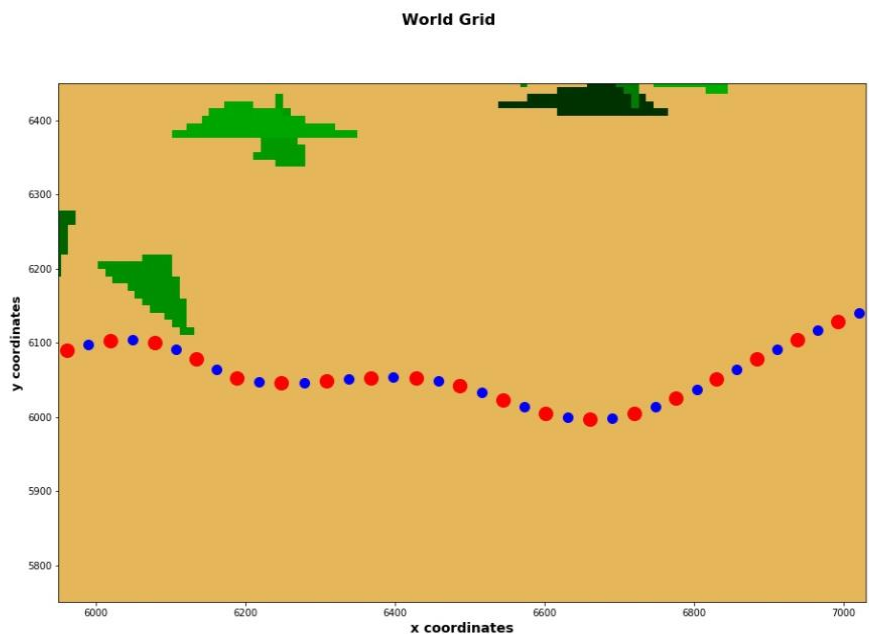

_Figure 23 This figure represents the training and the test dataset. Specifically, the blue dots represent the training dataset and the red dots represents the test dataset._

_Error of the model:_

The second form of evaluation is to calculate the median error between the set headings of the route snapshots and the recovered headings of the world grid images. In order to calculate this error, we followed the exact same way as we did in the Perfect Memory model.

### 4.4.4 Set up and Tuning of the Classifier

Another important aspect regarding the ANNs as classifiers implementation is to set up and tune the hyper-parameters. It is already known that there is no rule of thumb to select the structure and the hyper-parameters of a ANN as it depends on the problem, the complexity of the data and many other factors. The way that this problem can be approached is by brute force testing between different structures and hyper-parameters. An efficient way that is being used in machine learning problems is grid search. Grid search is an exhaustive search over a specified combination of parameters. Each of this combination of parameters corresponds to a specific model. The aim of grid search is to evaluate the performances of all of these possible models and find the most optimal combination of parameters.

In order to apply this technique in our model, we set all the parameters that we want to test and we run a grid search to find the best combination for our MLP classifier. Using this technique, we are able to identify the optimal structure (number of neurons / number of hidden layers) as well as other important hyper-parameters as the activation functions, the weight solvers and the learning rate for every training dataset.

### 4.4.5 Test MLP on a small part of the route

In the previous section we demonstrated the idea of "ANNs as classifiers" approach using only one snapshot to train the MLP and recover the heading of the closest world grid image. The next step is to test the performance of the MLP for a bigger part of the route.

We decided to test the MLP on the 20 first snapshots of the training route. After generating the training and the test dataset from these snapshots we had to tune our ANN. Using the Grid search approach described above we managed to find out the best tuning for the specific dataset.

According to the results of the Grid search (Appendix 7.4) the MLP should have two hidden layers with 20 neurons each, an identity activation function and a "lbfgs" weight solver. Looking at the Grid search results, we observe that "lbfgs" is the optimal solver for the 5 first models. This result confirms the statement that for small datasets, 'lbfgs' performs better, as in this case it was also used a small training dataset. (scikit-learn.org)

The next step was to fit the training data in the tuned MLP classifier and test its performance on the test dataset. Looking at the classification report (figure 24) we can conclude that our MLP managed to score accurate predictions on both "off" and "on" route testing data.

**Classification Report**

|          | precision | recall | f1-score | support |
|----------|-----------|--------|----------|---------|
| 0        | 1.00      | 1.00   | 1.00     | 10      |
| 1        | 1.00      | 1.00   | 1.00     | 10      |
| avg / total | 1.00   | 1.00   | 1.00     | 20      |

*Figure 24 This figure represents the Classification Report of MLP's performance on the test dataset.*

According to the classification report, our MLP had a 100% accuracy on both positive and negative testing data. Although this accuracy is difficult to be achieved, in this case it was used a small dataset and the training and testing data were very similar.

The second level of evaluation of our MLP model was to test its ability to recover the headings of the closest world grid images along the training route. Using the place recognition system, we managed to identify those images and test our MLP.

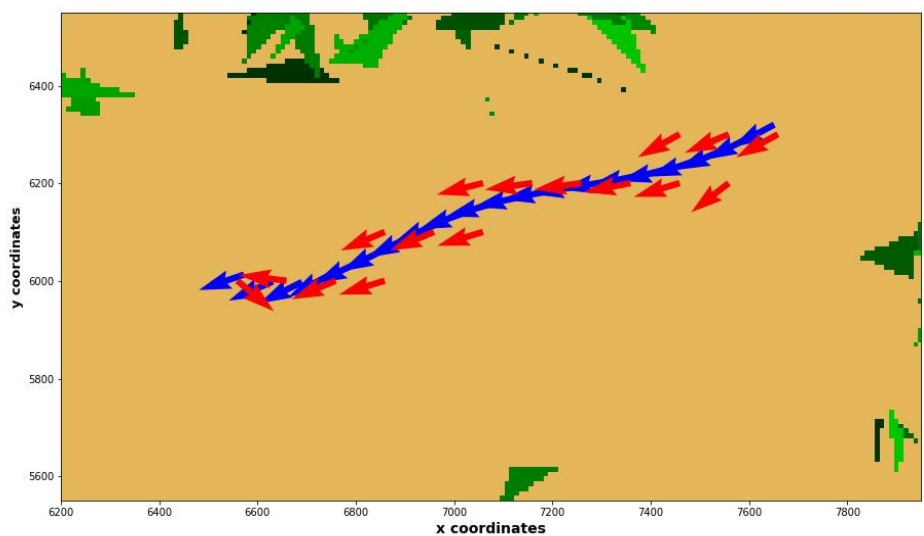

*Figure 25 This figure represents the Heading diagram between the 20 first set headings of the route snapshots and the recovered headings of the closest world images. The blue arrows represent the headings of the route and the red arrows represent the recovered headings.*

According to the produced heading diagram in figure 25, our model managed to recover the majority of the headings successfully. Specifically, the overall median error of the model along the 20 first snapshots of the route was $\approx 4^{o}$.

Overall, the MLP classifier managed to produce a really good performance for a small part of the route. The final step is to test the MLP along the whole training route and analyse the results of its performance.

### 4.4.6 Investigate the performance of the ANNs in the whole route

After examining the behavior of the MLP in a small part of the route we investigate how it can perform in the whole route. Following the same procedure, we collected the positive / negative training and testing data along the route. Subsequently, using grid search we found the best tuning and structure of the MLP for this new dataset.

According to the Grid search results, the optimal tuning for the MLP classifiers when using the whole dataset is 1 hidden layer with 100 neurons, "relu" activation function and "adam" weights solver. After tuning the MLP, the training data was fed into the model and we evaluated its performance. The first level of evaluation was testing the MLP on the test dataset.

**Classification Report**

|              | precision | recall | f1-score | support |
|--------------|-----------|--------|----------|---------|
| 0            | 0.96      | 0.93   | 0.95     | 46      |
| 1            | 0.94      | 0.96   | 0.95     | 46      |
| avg / total  | 0.95      | 0.95   | 0.95     | 92      |

*Figure 26 This figure represents the Classification report of the MLP's performance on the whole dataset.*

As shown in the Classification report, the MLP produced an accuracy of 95% on the whole route. Specifically, it managed to predict 94% of the "on route" testing views and 96% of the "off route" testing views. Although, the high accuracy of the classifier on the test dataset, it did not manage to produce a respectively good performance in the second level of the model's evaluation, the heading error. Specifically, it produced a median error of $\approx 45°$.

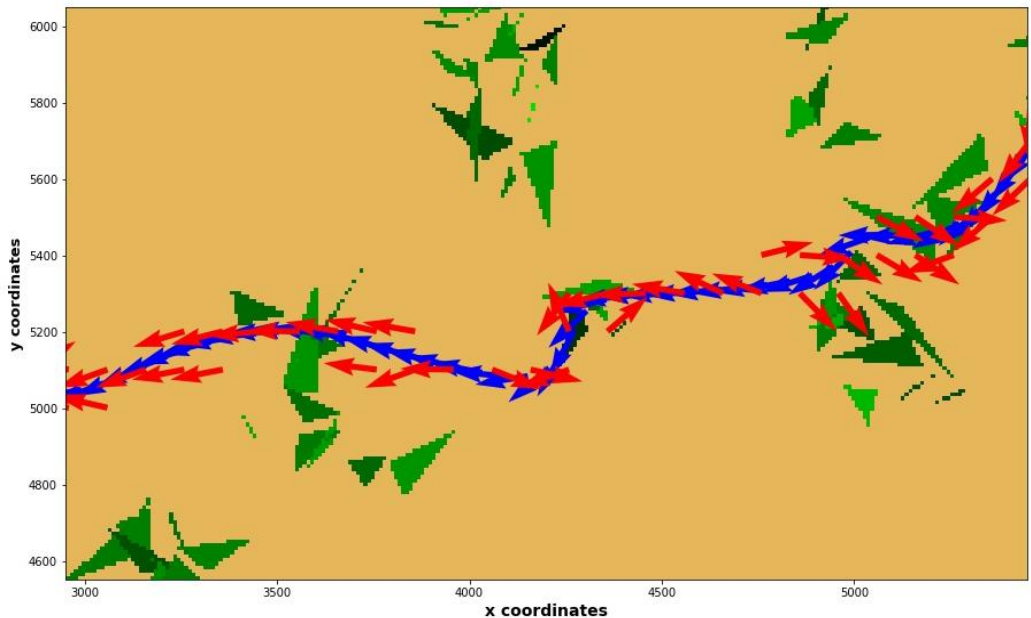

*Figure 27 This figure represents the heading diagram between the route and the recovered world grid headings. The blue arrows represent the route headings and the red arrows represent the recovered world grid headings.*

As shown in figure 27, our model did not manage to recover all the headings of the world grid. An interesting observation is that the MLP manages to recover more accurately the headings in the straight parts of the route. On the other hand, the big problem is being observed in the corners where it entirely fails to recover the correct headings.

In order to confirm this assumption, we had to compare the performance of the MLP with the Perfect Memory model on the same world grid images. If the Perfect Memory manages to recover the headings of the corner means that it is an actual problem of the "ANNs as classifiers" approach and needs extra investigation. Otherwise, it means that this problem is caused due to the possible problems that we discussed in the baselines methods sections (possible tussocks that blocks the visual field of the agent or the fact that the environment may look similar from different locations) and ANNs approach is not responsible for it.

Below it is depicted the heading diagram for the same world grid images, but this time using the Perfect Memory model to recover the headings

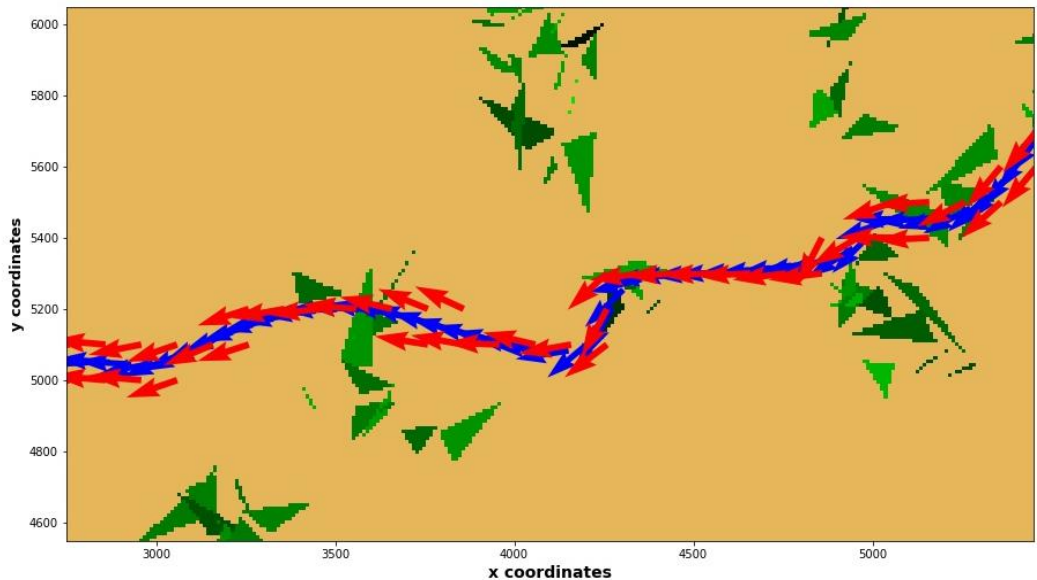

*Figure 28 This figure represents the heading diagram between the route and the recovered world grid images headings using Perfect Memory model. The blue arrows represent the route headings and the red arrows represent the recovered world grid headings.*

According to the heading diagram of the Perfect Memory model (figure 28), it manages to precisely recover all the headings of the world grid images without having problems in the corners of the route. That means that the weakness of recovering the headings in the corners of the route is a vulnerability of the "ANNs as classifiers" approach.

In the next section of the methodology chapter of the report we will analyse the behaviour of the "ANNs" as classifiers approach on recovering the headings in the corners of the route and describe the possible reason for this problem.

### 4.4.7 Investigate the performance of the ANNs in the corners of the route

Investigating the performance of the route during the whole training route helped us understand that our approach has a difficulty in recovering the headings in the corners of the route. This section will focus on the investigation for the possible reason regarding this behavior.

In order to train the MLP Classifier we use both positive and negative views generated from the snapshots along the training route. However, in order to train the Classifier correctly and produce accurate results we should use equally distributed training views from all along the training route.

Although, we collect views every six snapshots, the views along the route have many differences. Specifically, the rate of change in the differences of the images in the corners of the route is much bigger than the straight parts of the route. This is happening because the snapshots in the straight parts of the route are more similar with each other. This assumption can be confirmed by looking the next two diagrams. The first diagram depicts the collected snapshots from a specific part of the route and the second one depicts how the rms error varies between them.

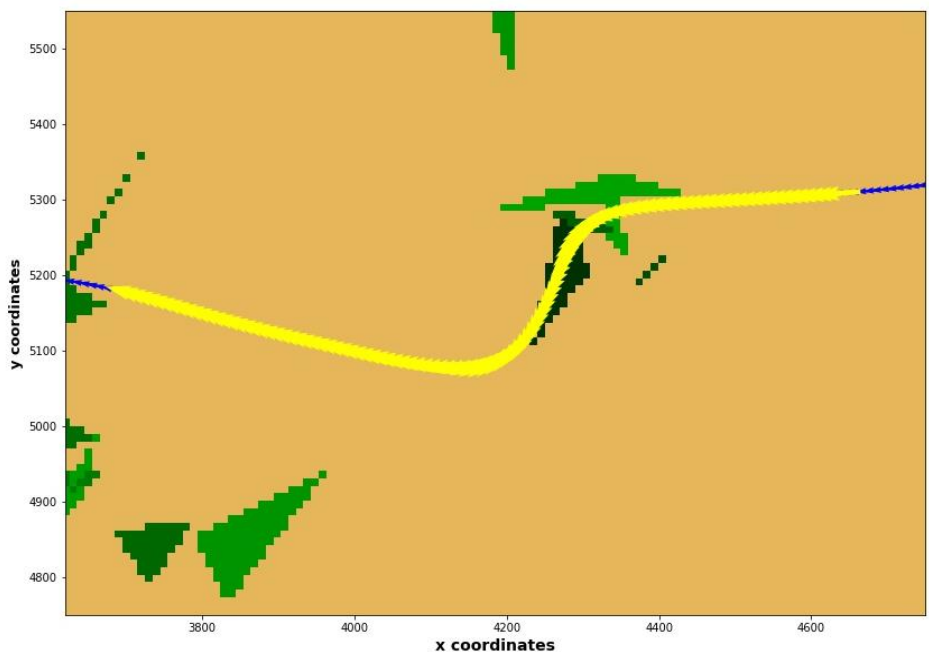

*Figure 29 This figure represents the selected part of the route that we decided to explore the rate of change in the differences of the images.*

The diagram in figure 29 represents the part of the route that we decided to examine. The reason that we decided to choose this part of the route is because it consists of two straight parts and two corners. Calculating the rms error between the successive snapshots we produced the next diagram, which depicts the rate of change of the differences of the images along this part of the route.

**Rate of change of the Differences**

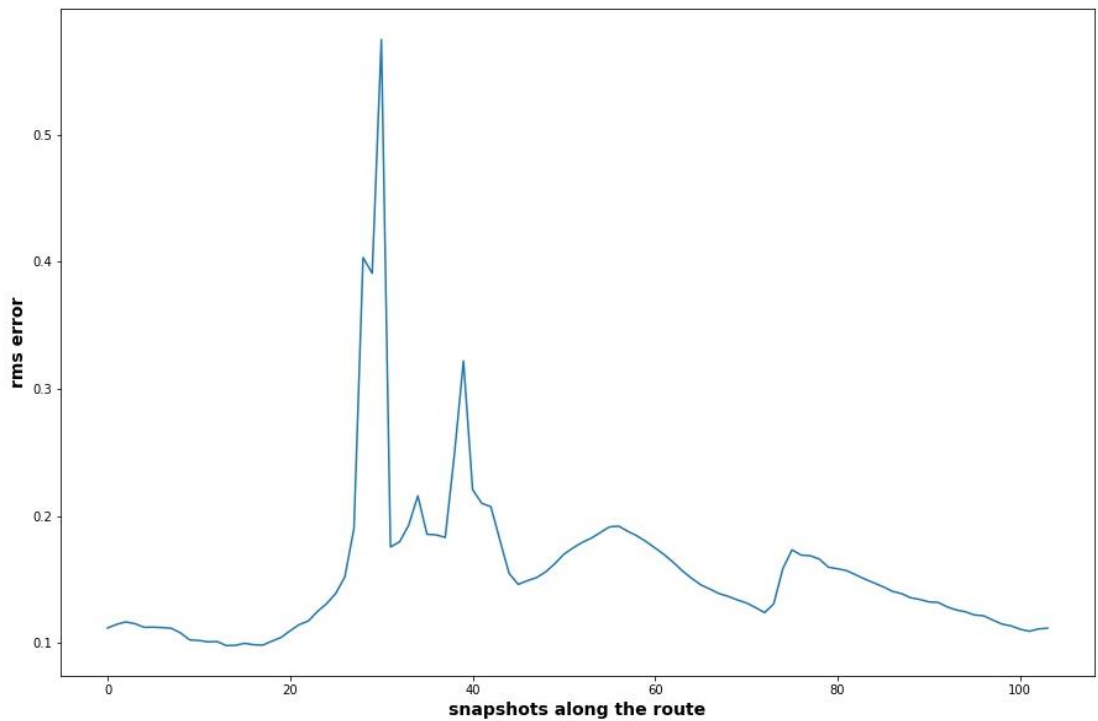

*Figure 30 This figure represents the rate of change of the rms error between the successive snapshots along the specified part of the training route*

According to the diagram in figure 30, the statements that the rate of change of the differences of the snapshots differs for every part of the route is being confirmed. Specifically, the rms error values which corresponds to the straight parts of the route are lower than the rms error values which corresponds to the corners of the route.

 Subsequently, that means that our model takes more training views similar to the straight parts of the route rather than the corners. This is a strong evidence why it performs better in the straight parts of the route rather than the corners as we do not used balanced training data.
In order to solve this problem, we approached to increase the training data in the corners of the training route.

Specifically, we decided to test it on the same part of the route that we did the investigation regarding the rate of change of the differences of the images. The reason that we selected to test this part of the route is because it has two successive corners and our model failed to recover the headings using the initial training data. In order to solve the problem, we collected more training images from this specific part of the route.

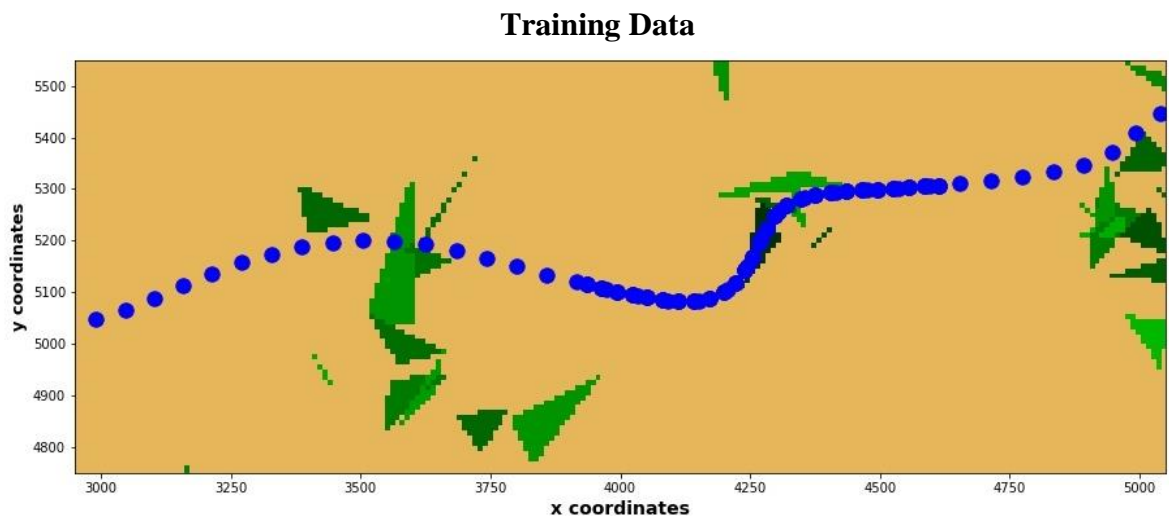

*Figure 31 This figure represents the training data used to feed our classifier. As you can see we used more training data from the corners of the route.*

A shown in figure 31 we collected snapshots with a frequency of 1/3 in the corners of the route and with 1/6 in the straight parts of the route. The next step was to tune our classifier for this new dataset (appendix 7.6) and evaluate its performance by plotting the recovered headings and compare them with the previous approach.

The first image in figure 32 depicts the heading diagram for the specific part of the route using equally distributed training data while the second image depicts the heading diagram produced using more training data in the corners. Comparing the two diagrams we can easily conclude that adding more training data in the corners improves the performance of the ANN for this part of the route. Specifically, in the first case it managed to recover 3/9 headings while with the second approach it managed to recover 7/9 headings. This result confirms our assumption that the reason for our model's weakness to recover the heading in the corners of the route was the lack of training views.

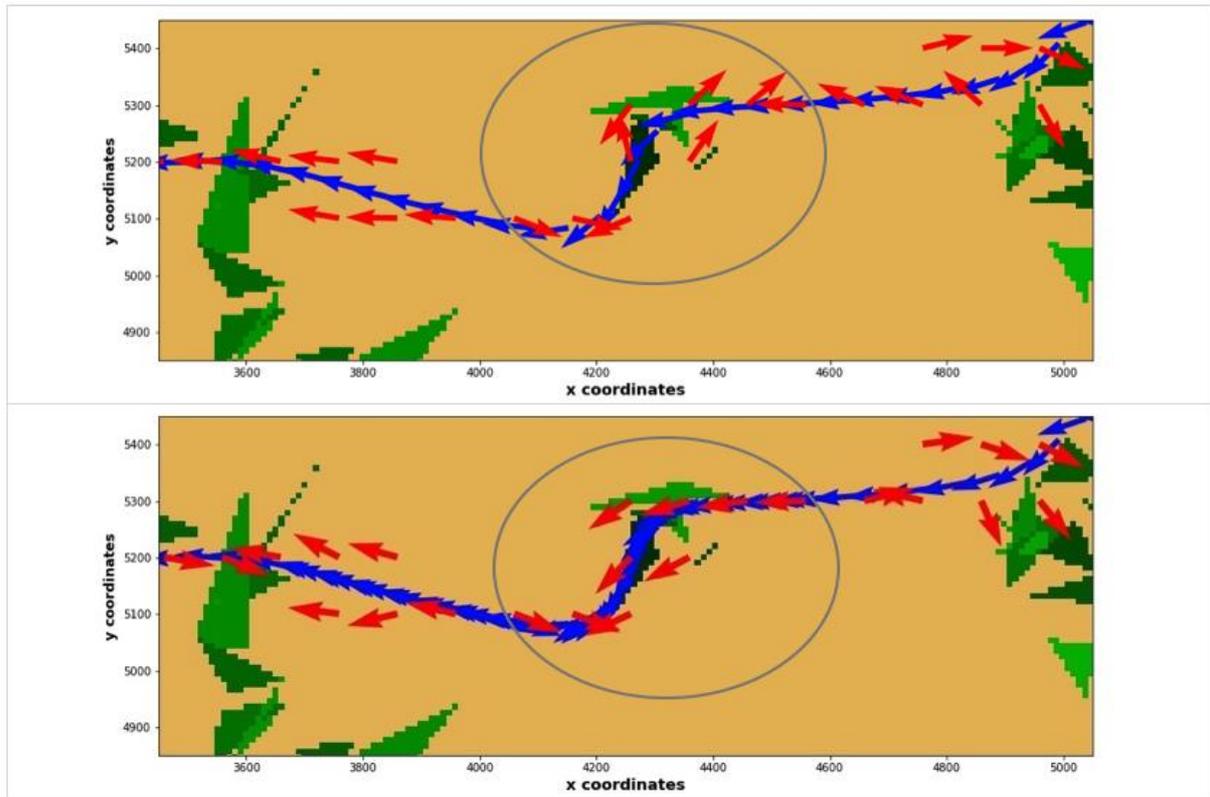

*Figure 32 This figure represents two heading diagrams. The first image represents the recovered headings using equally distributed training data. The second image represents the recovered heading using more data in the corners. The blue cycles represent the part of the route that we added the extra data.*

Although we managed to find a way to improve the performance of our model in the corners of the route we observed that there are other parts of the route that our model fails to recover the headings as well. The next and final part of the "ANNs as classifiers" section will focus on examining such cases by exploring the confidence values of the classifier and their distribution.

### 4.4.7 Investigate the confidence values of the ANNs

ANNs as classifiers approach produced some reasonable results in recovering the headings of the route. After understanding why it fails to recover the headings in the corners of the route, we will analyze what happens in the other parts of the route that it does not manage to recover the correct headings.

### *4.4.7.1 Confidence values and training time*

In order to start this investigation, we will plot and observe the confidence values diagrams of the recovered world grid images. In the figure below, they are depicted the confidence values diagrams from a correct recovered heading for different training time.

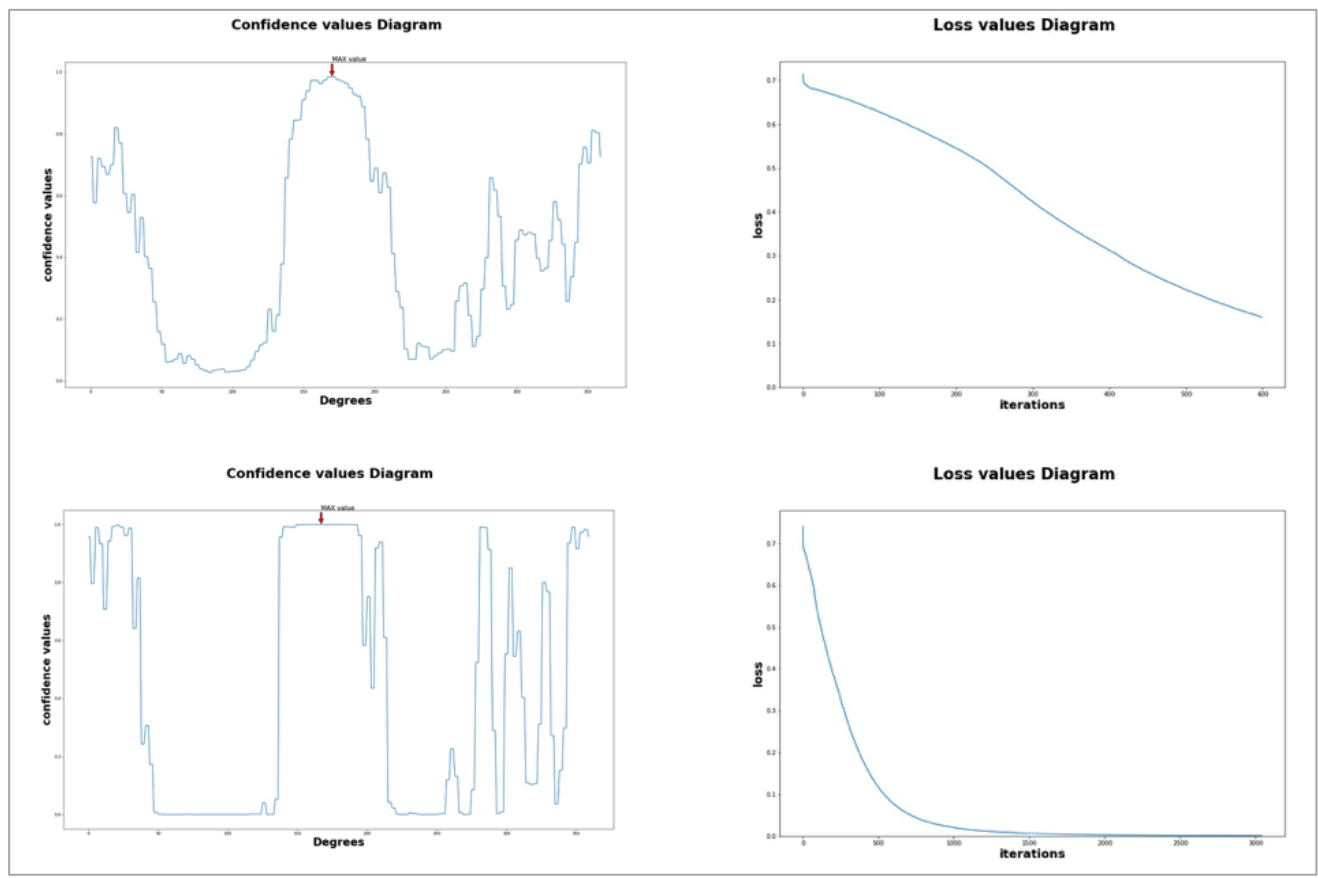

*Figure 33 This figure represents the confidence values diagrams for different training time. According to this diagram more training time implies to more biased models.*

The first diagram represent the confidence values diagram for 600 iterations while the second diagram represents the confidence values diagram for 3000 iterations. Observing the diagrams, we conclude that there is a trade of between the training time and biasing the model. Specifically, for a small training time (600 iterations) where loss values of the ANN are bigger, the distribution of the confidence values diagram are sharper presenting a distinct maximum confidence value. On the other hand, for a longer training time (3000 iterations) where the model converges to a minimum loss value, the distribution of the confidence values is bigger and it does not shape a distinct maximum value.

Although presenting a distinct maximum seems better in terms that it would be easier for our model to identify and recover the correct heading, after evaluating its performance for both these cases we found out that for 600 iterations it produced a really poor performance. Specifically, it produced a median error in the heading recovering of 110 degrees. On the other hand, when we let our model to converge in a minimum loss value (3000 iterations) its performance has a dramatically improvement as it produced an overall median error in heading recovering of 40 degrees.

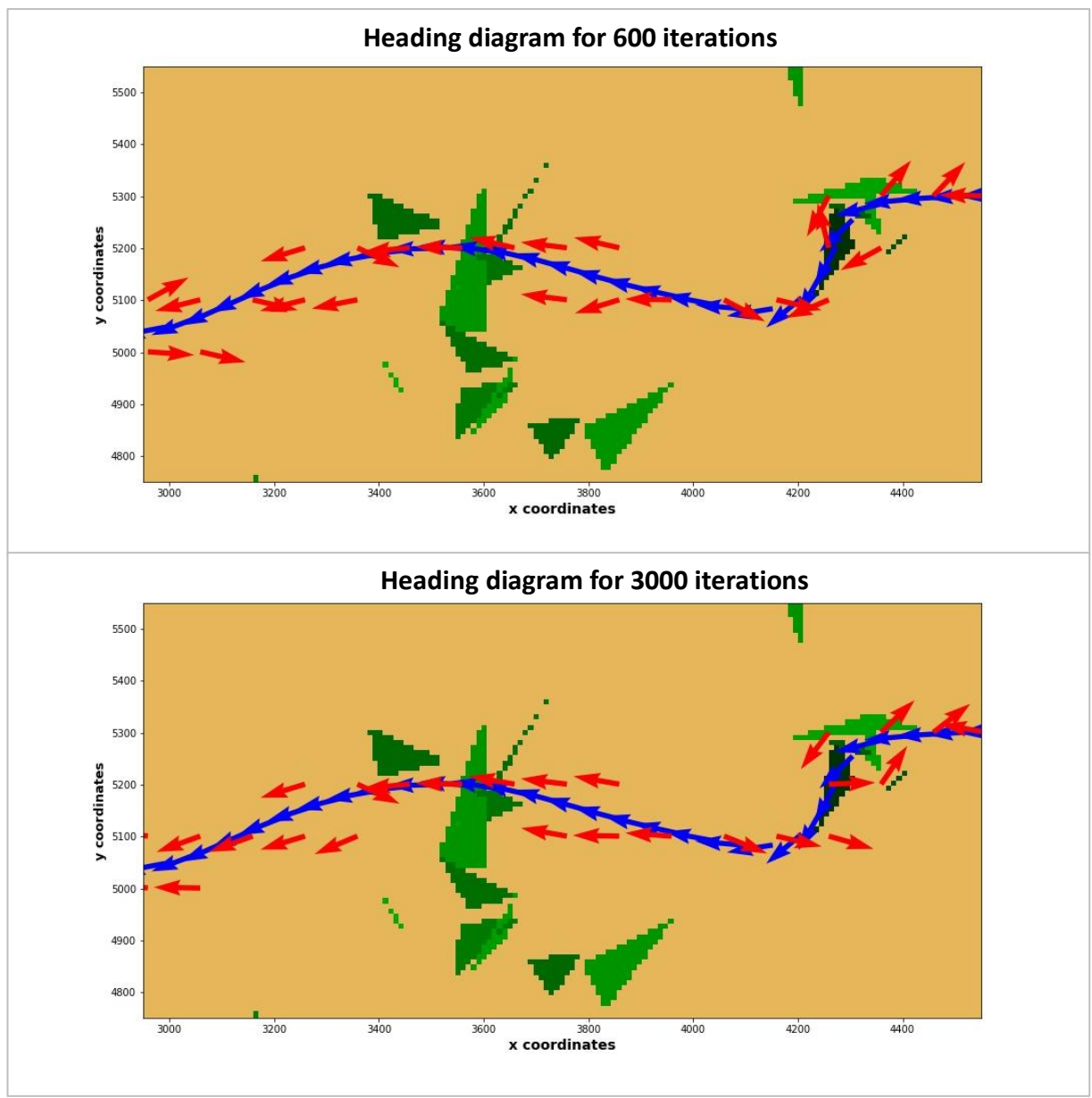

*Figure 34 This figure represents the Heading diagram for a training time of 600 and 3000 iterations respectively. According to this diagram the model managed to perform better for a longer training time as it managed to recover more headings.*

This can also be confirmed by figure 34 which depicts the heading diagram for both these cases. According to the heading diagrams our model performed better for a longer training time as it managed to recover much more headings. On the other hand, for a training time of 600 iterations it produced poor performance and it did not manage to recover much of the headings.

### 4.4.7.2 Confidence values and range of scanning

The next step after examining and understanding the relationship between the distribution of confidence values and the training time of the neural network was to understand why our model fails to recover some headings in the straight parts of the route and how the distribution of the confidence values is correlated with this issue.

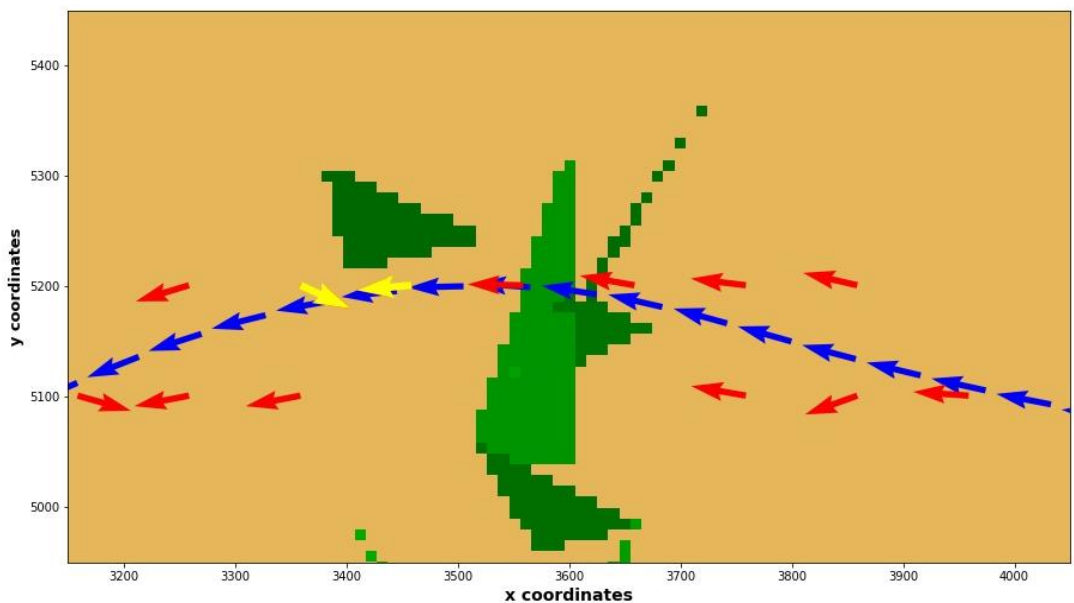

*Figure 35 This figure represents the headings that we will investigate their confidence values diagrams with yellow colour. We selected to investigate these two headings as they are both located in a straight part of the route. Headings are derived from ANN trained for 3000 iterations*

In order to do this, we investigate a case in which neural network does not manage to recover the heading in a straight part of the route (figure 35). In order to investigate this case, we decided to plot the confidence values diagrams between a world grid image which managed to recover its heading correctly and a world grid images which did not manage to recover its heading at all.

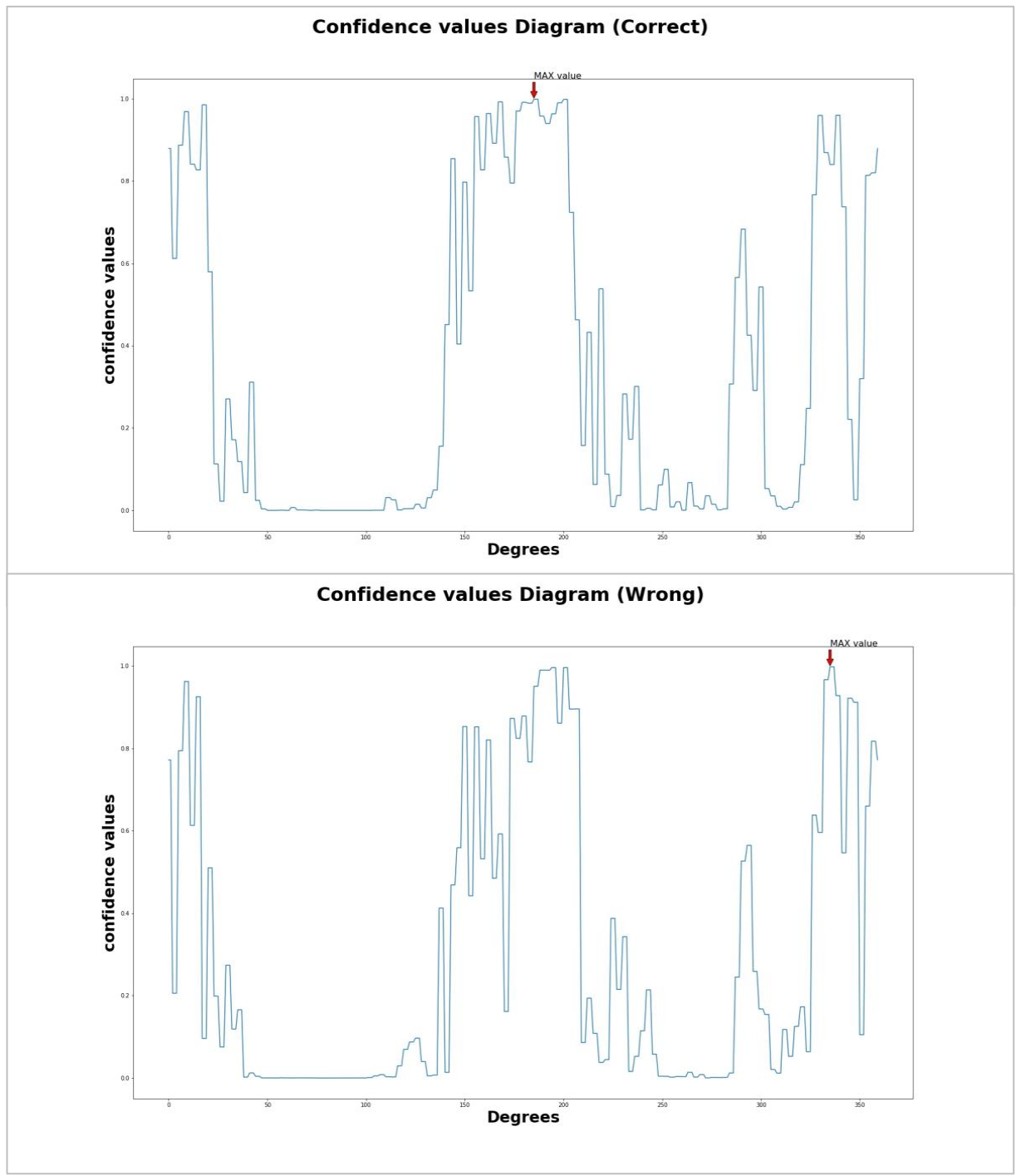

*Figure 36 This figure represents the confidence value diagrams of the two world grid images. The first diagram is the confidence values diagram of the world grid image which recovered the heading correctly while the second diagram represents the confidence values diagram of the world grid image which did not manage to recover its heading.*

According to the confidence values diagrams of the selected world grids images (figure 36) we can see that both diagrams managed to produce some high confidence values around the area of the correct heading (180º). However, they also produced some high confidence values for some other rotations of the image. If we observe the diagrams closely, we can discern that those rotations correspond to the opposite direction of the heading of the agent's route. A reasonable explanation for this could be found in the way that the training data was collected. Specifically, as it was also mentioned in the "ANNs as classifiers Implementation" section the way that we collected our training data was by using the snapshots as the positive training data and the rotated versions of those snapshots for the negative ones. In order to produce the negative training data, we rotated the snapshots 60º right and left from the route alternately. However, this way of collecting our training data did not provide any information about the views that have the opposite direction of the direction of the route. Taking into account that our ANN had no information about those views as well as the uniformity of the environment we can easily understand why it produced high accuracy values for these rotations of the snapshots.

Comparing the two confidence value diagrams we can conclude that the main reason that our model did not manage to recover the second world grid image is due to the large distribution of the high confident values for rotations which corresponds to the opposite direction from the direction of the route.

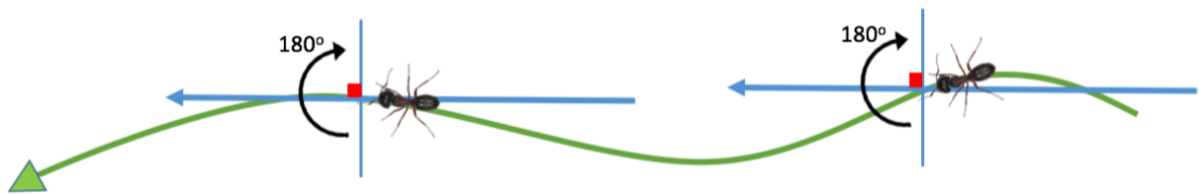

*Figure 37 This figure represents the proposed scanning of the agent in order to get rid of the high confidence values of the model which are produced for the rotations which corresponds to the opposite direction of the route. As you can see in the diagram the agent scans over 180º of the forward facing the general direction of the route. The green arrow represents the route and the blue arrows represents the general direction of the route.*

A way to get rid of these inaccurate predictions is to limit the search of the agent only in the forward facing direction of the route as shown in figure 37. Instead of doing a full rotation of 360º and recover the heading according to the highest confidence view we will limit this rotational search over 180º over the forward facing direction of the route. Specifically, we decided to limit this rotation over the general forward facing direction of the route. The reason that we selected to pick one stable direction for the whole route and do not do this search using as reference the direction of the closest snapshot's heading is that in this case the model would be too biased.

Following this approach, we will be able to discard the views of the opposite direction of the route and eliminate the inaccurate recovered headings.

After applying this technique, the model's overall performance significantly increased. Specifically, the overall median error decreased from 40º to 15º. Looking at the new heading diagram in figure 38 we can see that the model managed to recovered both of the headings that we examined before accurately.

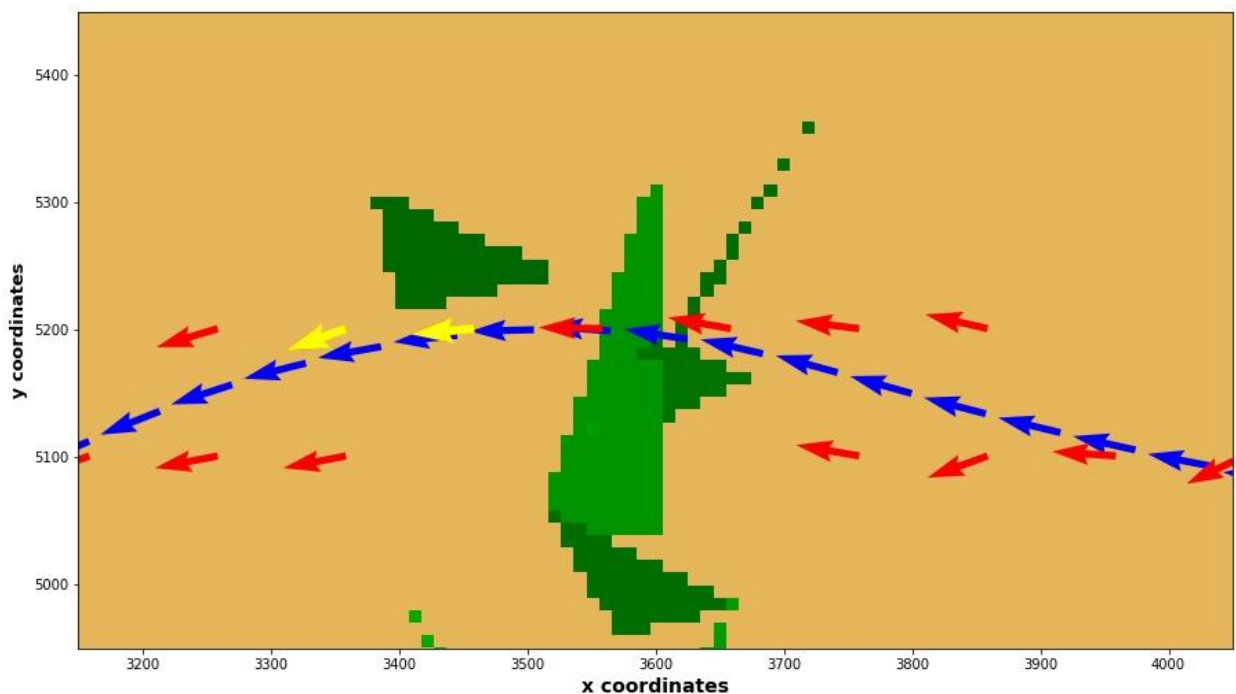

*Figure 38 This figure represents the heading diagram using the 180º rotation of forward facing direction to the route. According to this diagram both of the examined images managed to recover their headings to face the route.*

Finally, as we can see in the new confidence values diagram of the world grid image that it had not managed to recover its heading, we will see that the distribution of the high confidence values corresponding to the opposite to the route's direction views have been disappeared (figure 39) . On the other hand, the model identifies high confidence values only over the actual location of the direction of the route and manages to precisely recover the correct heading.

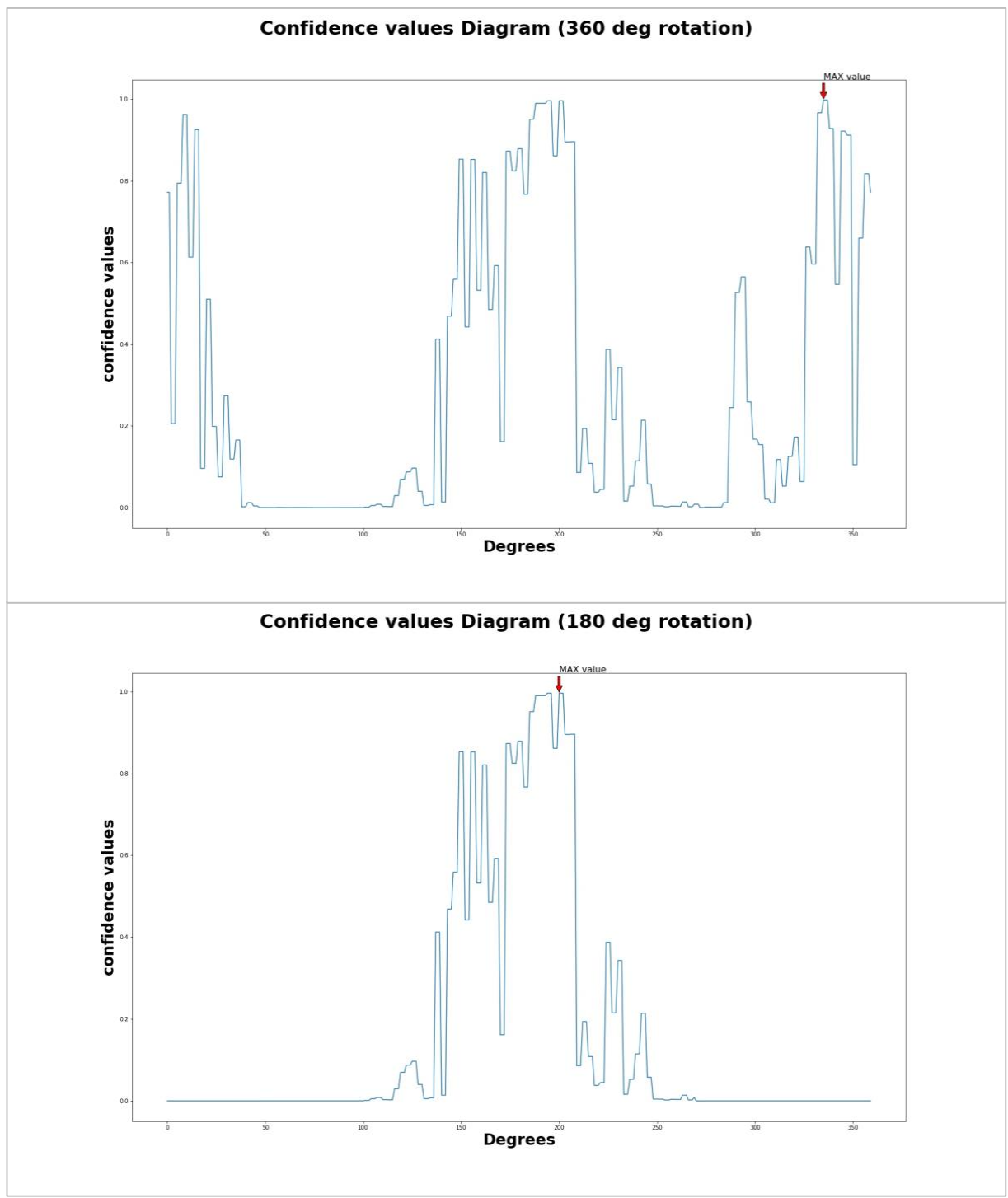

*Figure 39 This figure represents the confidence values diagram for the world grid image with coordinates (x, y) = (3359, 5201). The first diagram is the diagram produced following a 360° rotation. This diagram presents high confident values for the part of the rotation which corresponds to the opposite direction of the direction of the route and it does not manage to recover the correct heading. On the other hand, the second diagram is the confidence values diagram of the same world grid image but this time it was produced following a 180° rotation of forward facing direction of the route. This diagram presents high confident values only on the part of the route that corresponds to the direction of the route. Following this approach, we discard the views that corresponds to the opposite direction of the direction of the route and as a result we get rid of the noise that they produced to the diagram.*

Another interesting observation was that further decreasing the field of rotational search implied to even better performance of the model. Specifically, reducing the rotational scanning to 90º over the general forward facing direction of the route decreased the overall median error to 10º. However, reducing the rotational search in such a limited part of the route would biased the model a lot and conclude to unrealistic results.

The investigation of the confidence values and their distribution helped us to understand a basic limitation of the initial model and the collection of the training data. However, discarding the unseen views by changing the rotation of the agent helped our model to improve its performance and recovered much more headings.

# 6. Conclusion and Future Work

Reaching the conclusion section, it could be stated that the aim of this project was achieved. Throughout this project we managed to implement "Perfect Memory" and "ANNs as classifiers", two different approaches that both managed to successfully replicate the visual navigational behavior of the ants. Analyzing the performance of both these implementations helped us determine what approaches perform well, what approaches do not provide satisfying results and why. This investigation drove the research towards some very compelling directions providing outcomes that helped us overcome the models vulnerabilities and improve their performances.

Using Perfect Memory as baseline method, we managed to investigate what Preprocessing and Frequency of training data we should use in our advanced ANNs approach. During this investigation we came up with the expected outcome that mid-resolution images are better suited for navigation. This result is also supported by previous research made by Dr. Graham and Dr. Philippides as mentioned in the background section. Subsequently, we focused on implementing the "ANNs as classifiers" approach and evaluating its performance. We found that this approach performed better when applied to a small part rather than the entirety of the route. Analysing the heading diagrams for the whole route, we observed that the main problem of this approach was recovering the headings at the corners of the route where the rate of change of the views is bigger. In order to overcome this problem, we decided to increase the number of the training data along these parts of the route which had a positive impact on the model's performance. The final part of this investigation was focused on the distribution of the confidence values and their impact on the model's performance. After analysing the relationship between the confidence values and the training time we concluded that the ANN performs better when provided with enough training time to converge to a loss minimum. Finally, we observed that the distribution of the confidence values of different world grid images was extremely high for the parts of the rotations corresponding to the opposite of the route's forward facing direction. The reason for this was the lack of training data regarding the backward part of the route. That caused the model to sometimes fail when recovering headings even though they were located in straight parts of the route.

This problem was overcome by suggesting that the agent should scan its environment in a way that discards the views corresponding to the opposite of the route's forward facing direction. After applying this approach our model's performance increased dramatically and provide us with an overall error of $\approx 17^\circ$.

In terms of further development there is a lot to be done. This project opens a few directions for extra investigation. A potential future research could be the investigation of more efficient selection of the training data in the corners of the route. According to our results, adding more training data in the corners of the route improves the model's performance. However, every corner has a different angle and as a result a different rate of change of the differences of its images. An efficient selection based on this fact could be helpful to further improve the performance of the model in these parts of the route. Another possible improvement of the project could be the use of different classifiers for different parts of the route. As it was also proved by our research, MLP classifier can perform better for a small part of the route. So, using different classifiers for different parts of the route could be a way to improve the performance of the model as each classifier would handle less training data. Finally, another interesting investigation would be to apply a feature selection technique to our training data. That means that instead of feeding the whole pictures as training data, we would extract only the interesting points of the images. All these proposed investigations have many potentials for improving the existing performance of the model and would generate new ideas and findings regarding the replication of the ants' visual navigation behavior using "ANNs as classifiers" approach.

As I approach the end of this project, I would like to state that I feel privileged to have the opportunity of focusing my research on the field of the biomimetic algorithms. My research regarding insect inspired algorithms and particularly the autonomous visual navigation of the ants inspired me to combine my previous academic knowledge from a range of scientific areas and implement different types of biomimetic algorithms which replicates the navigational behavior of the ants. Engaging with this project also helped me understand the basic principles of a research project and further develop the skills necessary to dive deeper into this field.

# 7. References

**[1]** Baddeley, B., Graham, P., Husbands, P. and Philippides, A. (2012). A Model of Ant Route Navigation Driven by Scene Familiarity. PLoS Computational Biology.

**[2]** Bcs.org. (2018). *Code of conduct | Membership | BCS - The Chartered Institute for IT*. [online] Available at: https://www.bcs.org/category/6030

**[3]** Cartwright B.A and Collett T.S (1983). Landmark learning in bees.

**[4]** Collett, T. S., Graham, P., Harris, R. A. and Hempel-De-Ibarra, N. 2006. Navigational memories in ants and bees: Memory retrieval when selecting and following routes. Adv. Stud Behav, 36, 123-172.

**[5]** Graham, P. and Philippides, A. (2017). Vision for navigation: What can we learn from ants? Arthropod Structure & Development, 46(5), pp.718-722.

**[6]** Lambrinos, D., Möller, R., Pfeifer, R. and Wehner, R. (1998). Landmark Navigation without Snapshots: The Average Landmark Vector Model.

**[7]** Land, M. and Nilsson, D. (2012). Animal eyes. Oxford: Oxford University Press.

**[8]** Lee, C. and Kim, D. (2017). Local Homing Navigation Based on the Moment Model for Landmark Distribution and Features. Sensors.

**[9]** Möller, R., Lambrinos, D., Pfeirer, R. and Wehner, R. (1998). Insect Strategies of Visual Homing in Mobile Robots

**[10]** Philippides A., Baddeley B., Husbands P., Graham P. (2012) How Can Embodiment Simplify the Problem of View-Based Navigation? In: Prescott T.J., Lepora N.F., Mura A., Verschure P.F.M.J. (eds) Biomimetic and Biohybrid Systems. Living Machines 2012. Lecture Notes in Computer Science, vol 7375. Springer, Berlin, Heidelberg

**[11]** Philippides, A. (2013). Visual route navigation in ants: A situated and embodied approach presentation.

**[12]** Philippides, A., Graham, P., Baddeley, B., Husbands, P. (2015). Using neural networks to understand the information that guides behaviour: a case study in visual navigation.

**[13]** Scikit-learn.org. *sklearn.neural_network.MLPClassifier — scikit-learn 0.20.3 documentation*. [online] Available at: https://scikit-learn.org/stable/modules/generated/sklearn.neural_network.MLPClassifier.html

**[14]** Wehner, R. and Wehner, S. 1986. Path Integration in Desert Ants - Approaching a Long Standing Puzzle in Insect Navigation. Monitore Zoologico Italiano-Italian Journal of Zoology, 20, 309-331.

**[15]** Wehner, R. and Srinivasan, M. 2003. Path integration in insects. In: The neurobiology of spatial behaviour (Ed. by K. Jeffrey, J.), pp. 9-30. Oxford: Oxford university press.

**[16]** Wehner, R. and Srinivasan, M. 2003. Path integration in insects. In: The neurobiology of spatial behaviour (Ed. by K. Jeffrey, J.), pp. 9-30. Oxford: Oxford university press.

**[17]** Wittlinger, M., Wehner, R. and Wolf, H., 2006. The ant odometer: stepping on stilts and stumps. Science, 312(5782), pp.1965-1967.

**[18]** Zanaty, E. (2012). Support Vector Machines (SVMs) versus Multilayer Perception (MLP) in data classification. *Egyptian Informatics Journal*, 13(3), pp.177-183.

# 7. Appendices

## 7.1 Brains on Board ethical statement

| Legal | General audience | Commentary |
|---|---|---|
| Robots are multi-use tools. Robots should not be designed solely or primarily to kill or harm humans, except in the interest of national security reasons. | Robots should not be designed as weapons, except for national security reasons. | BoB robots are primarily designed as basic research tools to better understand the autonomy of animals and artificial agents. BoB will work towards applications such as autonomous robots for agriculture and search and rescue. We recognise that other roboticists may re-use our technology and BoB is committed to advocate responsible use and do everything possible to discourage misuse. BoB will never get involved in any weapons related research or application. |
| Humans, not robots, are responsible agents. Robots should be designed, operated as far as is practicable to comply with existing laws & fundamental rights & freedoms, including privacy. | Robots should be designed and operated to comply with existing law, including privacy. | The robots designed and used within BoB will have bio-mimetic controllers based on bee brains. Because the controllers are simulated brains, their actions are not governed by a simple rule set that could incorporate privacy laws, and laws around safe drone operation. Therefore, BoB robots will be operated under close supervision of humans including an over-riding manual "off" switch. The BoB consortium is committed to additional safeguards for eventual applications, including manual override and additional "virtual limits", e.g. to ensure a safe distance to humans is enforced at all times. |
| Robots are products. They should be designed using processes which assure their safety and security. | Robots are products: as with other products, they should be designed to be safe and secure. | BoB is based on existing drone hardware that has already undergone safety certification. The developed bio-mimetic controllers will be made safer by an additional rule system as described above. The robots will also be secured from hacking by secure cryptographic protocols. Product prototypes will be extensively tested for their safe use. |
| Robots are manufactured artefacts. They should not be designed in a deceptive way to exploit vulnerable users; instead their machine nature should be transparent | Robots are manufactured artefacts: the illusion of emotions and intent should not be used to exploit vulnerable users. | The robots are not intended for any roles that include social interaction with humans. The nature of the technology will be fully transparent to technical personnel through Open Source development approaches. |
| The person with legal responsibility for a robot should be attributed. | It should be possible to find out who is responsible for any robot. | During research the ownership and responsibility of BoB researchers will be obvious through their physical presence. BoB supports the notion that future products should undergo a registration and licensing scheme. |

## 7.2 Frequency of training data Perfect Memory Results

Below you can find the diagrams produced by the frequency analysis of the training data using Perfect Memory model. The first diagram represents the Average error of the model vs Frequency of the training data, the second diagram represents the Median error of the model vs Frequency of the training data and the Final diagram is the corresponding Whisker Box.

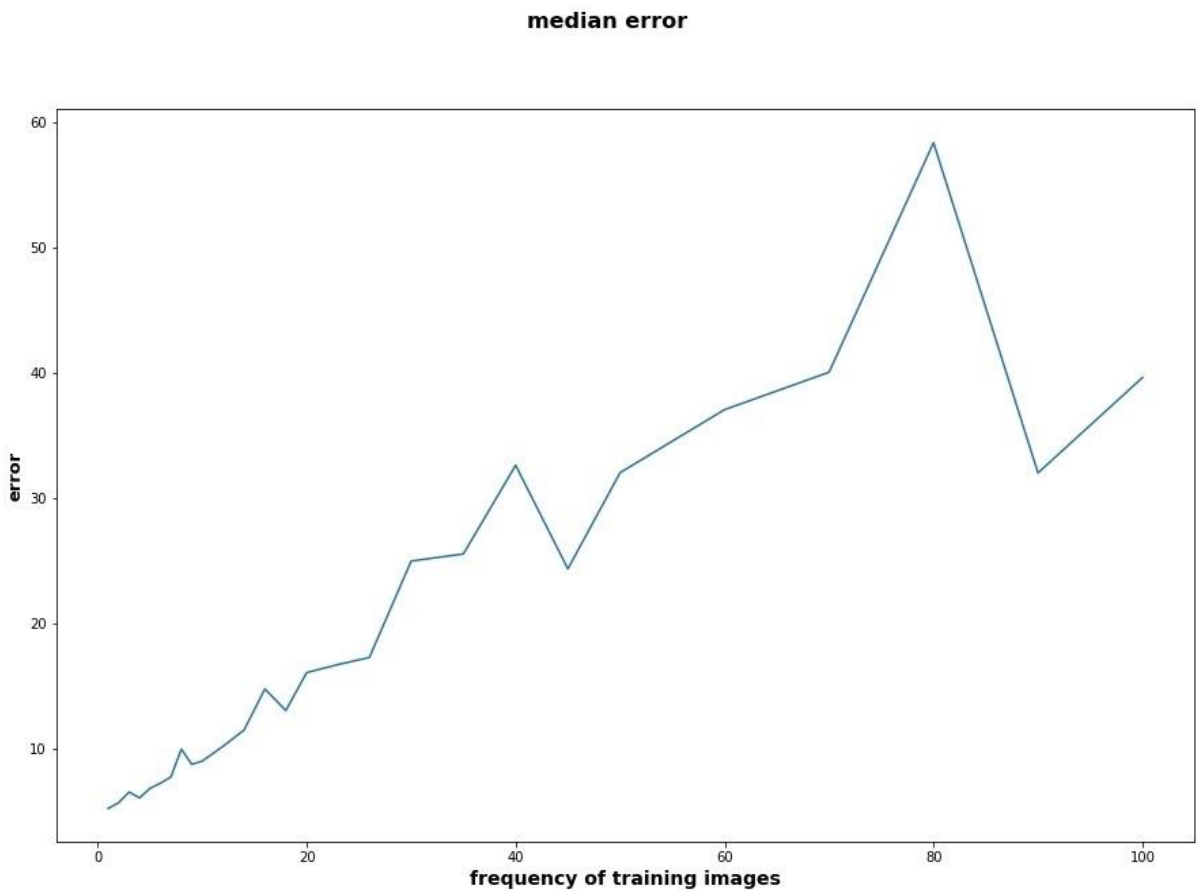

**Average error**

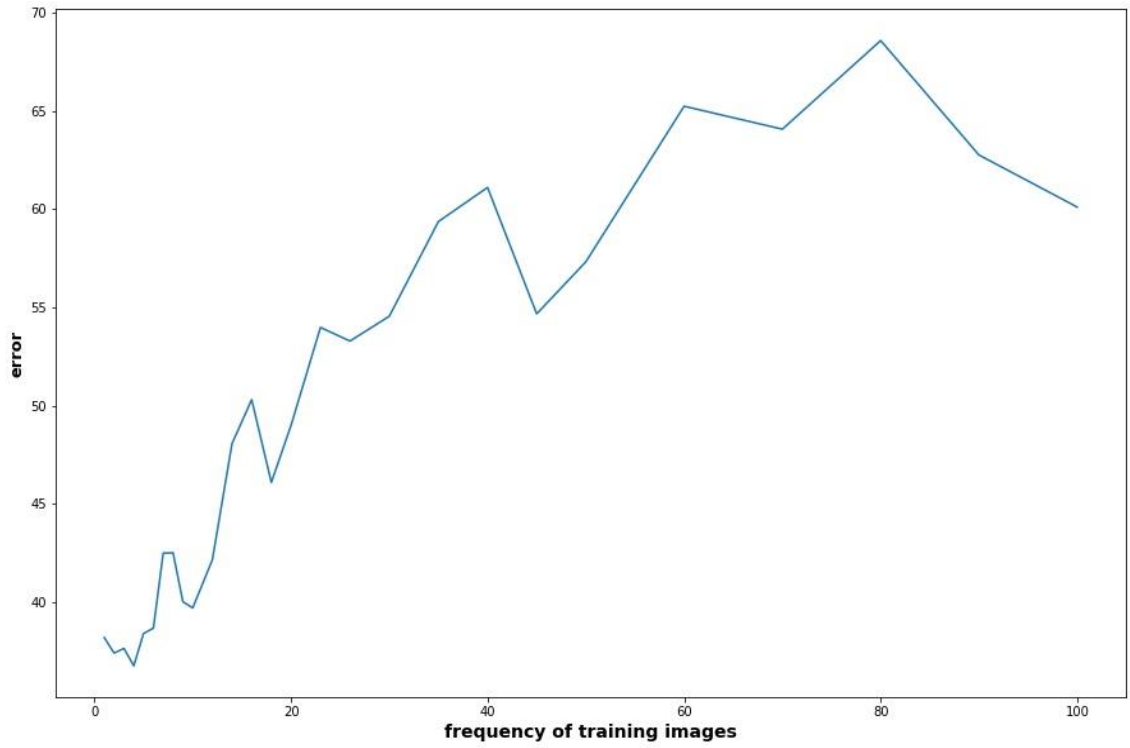

**Whisker box Diagram**

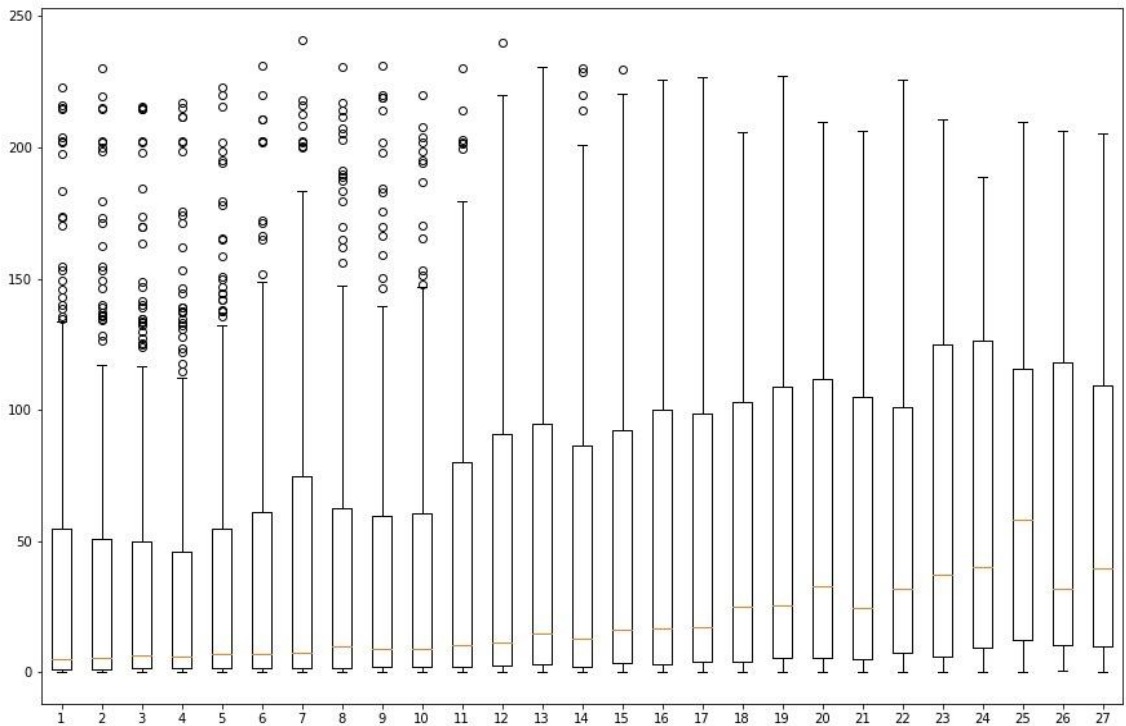

## 7.3 Preprocessing of training data Perfect Memory Results

Below you can find the diagrams produced by the Preprocessing analysis of the training data using Perfect Memory model. The first three diagrams depict the average, the median and the corresponding whisker box of the "error vs Preprocessing" of the model (with Gaussian smoothing) .

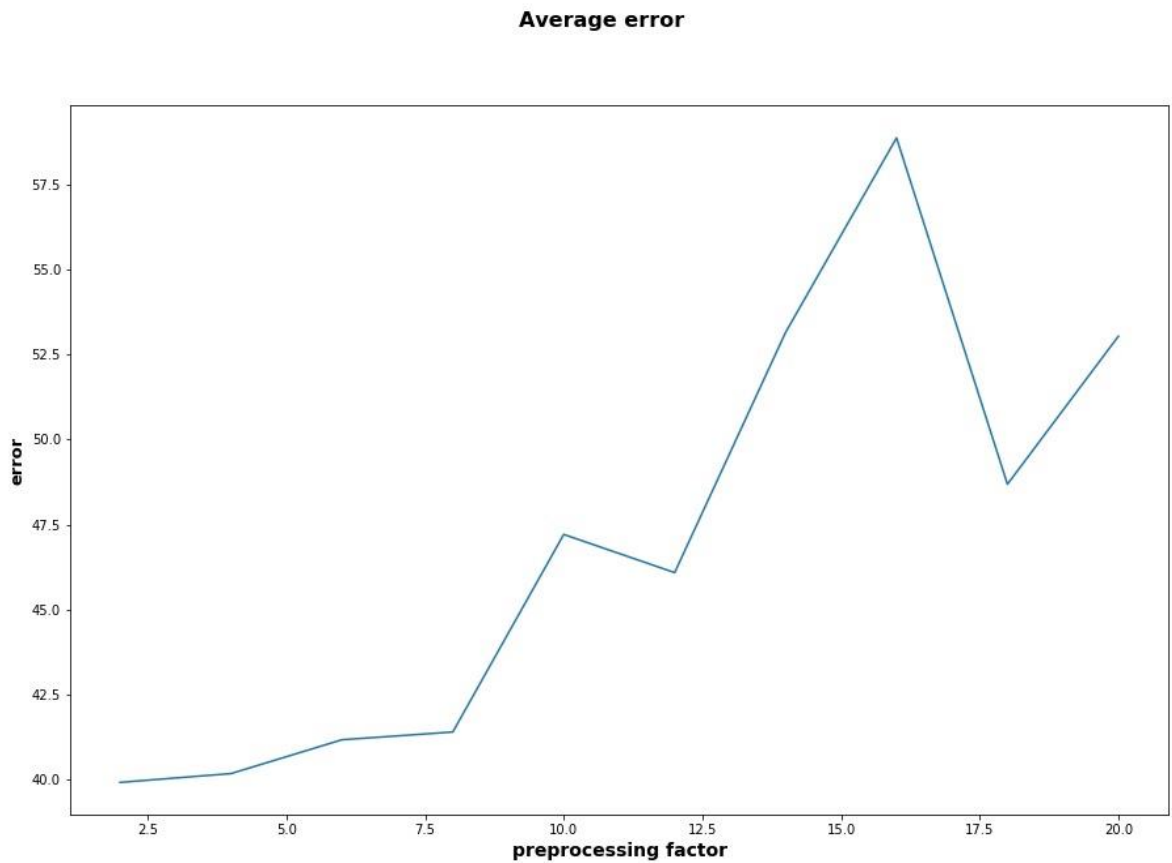

**median error**

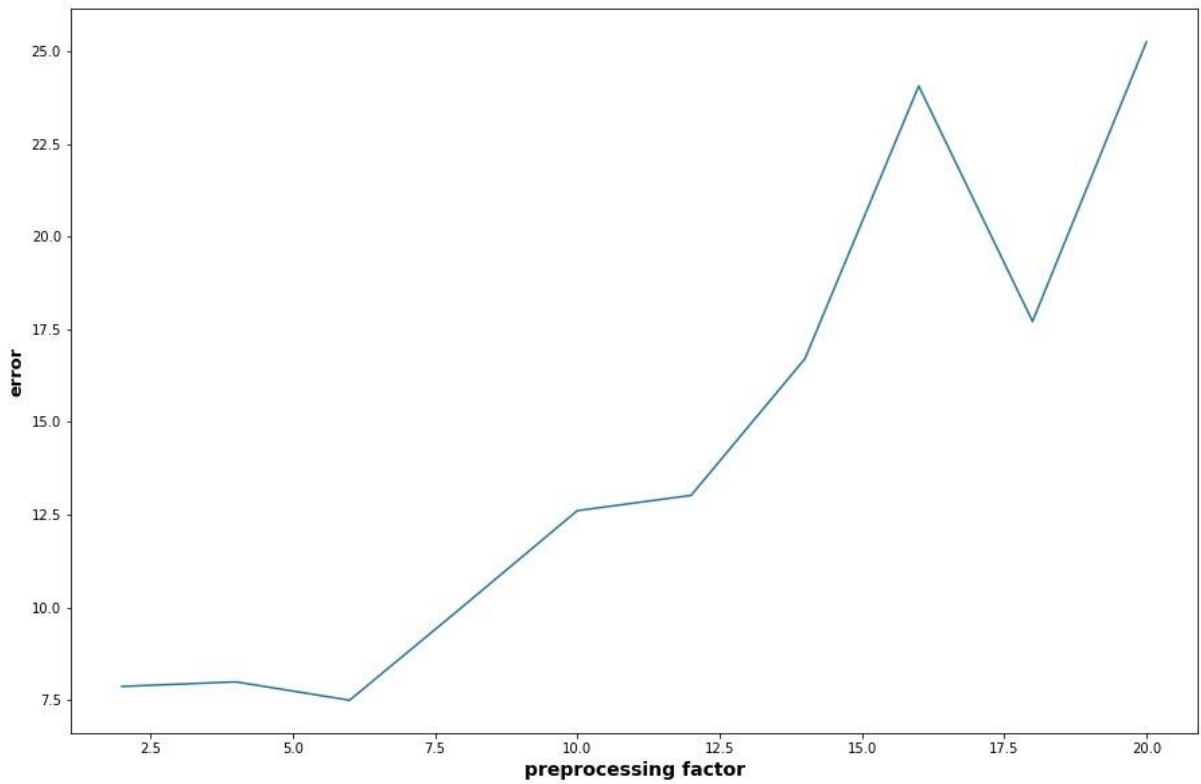

**Whisker box Diagram**

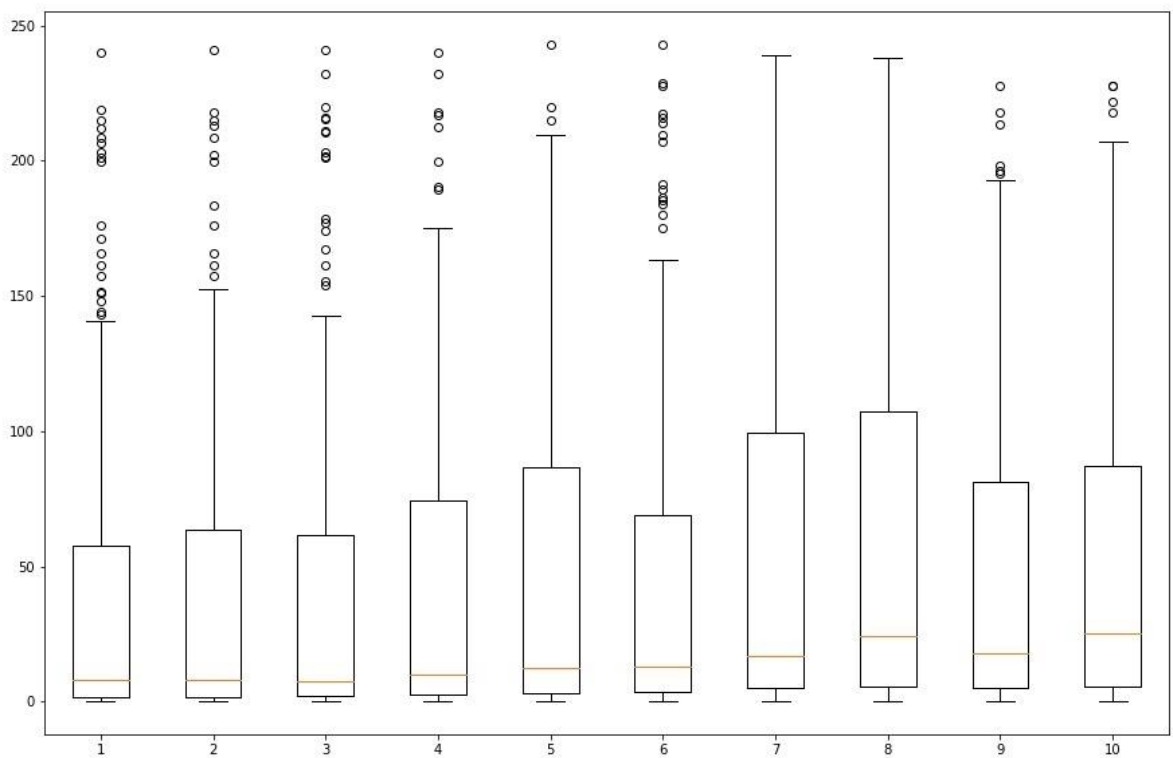

Below you can find the distinct values of the average / median error for the different resolution reductions with and without Gaussian smoothing.

### *Without Gaussian smoothing:*

```
In [29]:  avg_error

Out[29]:  [39.93135454545456,
           40.18992727272728,
           41.183154545454535,
           41.41128181818182,
           47.21370909090915,
           46.092309090909076,
           53.146509090909085,
           58.858454545454535,
           48.68729090909095,
           53.033518181818216]
```

```
In [30]:  median_error

Out[30]:  [7.864999999999995,
           7.989000000000185,
           7.493499999999997,
           10.035999999999987,
           12.605499999999992,
           13.018000000000015,
           16.7025,
           24.060000000000002,
           17.710000000000008,
           25.247000000000014]
```

### *With Gaussian smoothing:*

```
In [37]:  avg_errorGS

Out[37]:  [40.77516363636363,
           41.18444545454546,
           40.725363636363646,
           39.88190909090908,
           41.6775,
           44.523772727272714,
           49.822945454545454,
           52.424227272727265,
           47.9005636363636365,
           52.78776363636365]
```

```
In [38]:  median_errorGS

Out[38]:  [7.690000000000012,
           7.361000000000004,
           7.173999999999992,
           8.444500000000005,
           10.467500000000015,
           12.10199999999999,
           13.539000000000001,
           18.854,
           18.467000000000013,
           20.636500000000012]
```

## 7.4 Grid Search small part of the route

Below you can find the GridSearchCV results for the training of the first 20 snapshots of the route. The models have been placed in descending order according to their *mean_ test score* (mean validation score).

| | mean_test_score | mean_train_score | param_hidden_layer_sizes | param_activation | param_solver | param_tol | param_learning_rate_init |
|---|---|---|---|---|---|---|---|
| 206 | 0.850 | 1.000000 | (20, 20) | relu | lbfgs | 1e-05 | 0.0001 |
| 179 | 0.825 | 1.000000 | 20 | relu | lbfgs | 1e-05 | 0.0001 |
| 197 | 0.825 | 1.000000 | (20, 20) | relu | lbfgs | 1e-05 | 0.001 |
| 187 | 0.800 | 1.000000 | 20 | relu | lbfgs | 0.0001 | 1e-05 |
| 80 | 0.800 | 1.000000 | 20 | logistic | lbfgs | 1e-05 | 1e-05 |
| 1 | 0.800 | 1.000000 | 20 | identity | sgd | 0.0001 | 0.001 |
| 215 | 0.800 | 1.000000 | (20, 20) | relu | lbfgs | 1e-05 | 1e-05 |
| 188 | 0.800 | 1.000000 | 20 | relu | lbfgs | 1e-05 | 1e-05 |
| 191 | 0.800 | 1.000000 | (20, 20) | relu | sgd | 1e-05 | 0.001 |
| 17 | 0.775 | 1.000000 | 20 | identity | lbfgs | 1e-05 | 0.0001 |
| 170 | 0.775 | 1.000000 | 20 | relu | lbfgs | 1e-05 | 0.001 |
| 163 | 0.775 | 1.000000 | 20 | relu | sgd | 0.0001 | 0.001 |
| 190 | 0.775 | 1.000000 | (20, 20) | relu | sgd | 0.0001 | 0.001 |
| 71 | 0.775 | 1.000000 | 20 | logistic | lbfgs | 1e-05 | 0.0001 |
| 62 | 0.775 | 1.000000 | 20 | logistic | lbfgs | 1e-05 | 0.001 |
| 56 | 0.775 | 1.000000 | 20 | logistic | sgd | 1e-05 | 0.001 |
| 26 | 0.775 | 1.000000 | 20 | identity | lbfgs | 1e-05 | 1e-05 |
| 0 | 0.775 | 1.000000 | 20 | identity | sgd | 0.001 | 0.001 |
| 6 | 0.775 | 1.000000 | 20 | identity | lbfgs | 0.001 | 0.001 |
| 8 | 0.775 | 1.000000 | 20 | identity | lbfgs | 1e-05 | 0.001 |
| 79 | 0.750 | 1.000000 | 20 | logistic | lbfgs | 0.0001 | 1e-05 |
| 98 | 0.750 | 1.000000 | (20, 20) | logistic | lbfgs | 1e-05 | 0.0001 |
| 70 | 0.750 | 1.000000 | 20 | logistic | lbfgs | 0.0001 | 0.0001 |
| 7 | 0.750 | 1.000000 | 20 | identity | lbfgs | 0.0001 | 0.001 |
| 87 | 0.750 | 1.000000 | (20, 20) | logistic | lbfgs | 0.001 | 0.001 |
| 89 | 0.750 | 1.000000 | (20, 20) | logistic | lbfgs | 1e-05 | 0.001 |
| 169 | 0.750 | 1.000000 | 20 | relu | lbfgs | 0.0001 | 0.001 |
| 162 | 0.750 | 1.000000 | 20 | relu | sgd | 0.001 | 0.001 |
| 202 | 0.750 | 0.883333 | (20, 20) | relu | adam | 0.0001 | 0.0001 |
| 160 | 0.750 | 1.000000 | (20, 20) | tanh | lbfgs | 0.0001 | 1e-05 |
| ... | ... | ... | ... | ... | ... | ... | ... |

## 7.5 Grid Search whole route

Below you can find the GridSearchCV results for the whole training route. The models have been placed in descending order according to their *mean_ test score* (mean validation score).

| | mean_test_score | mean_train_score | param_hidden_layer_sizes | param_activation | param_solver | param_tol | param_learning_rate_init |
|---|---|---|---|---|---|---|---|
| 713 | 0.771739 | 0.875000 | 100 | relu | adam | 1e-06 | 1e-06 |
| 353 | 0.733696 | 0.887681 | 100 | logistic | adam | 1e-06 | 1e-06 |
| 303 | 0.722826 | 0.932971 | 100 | logistic | sgd | 1e-06 | 0.001 |
| 608 | 0.722826 | 0.954710 | (20, 20) | relu | adam | 1e-06 | 0.001 |
| 532 | 0.722826 | 0.920290 | 100 | tanh | adam | 1e-05 | 1e-06 |
| 354 | 0.722826 | 0.884058 | 100 | logistic | adam | 1e-07 | 1e-06 |
| 319 | 0.722826 | 0.869565 | 100 | logistic | sgd | 1e-07 | 0.0001 |
| 506 | 0.717391 | 1.000000 | 100 | tanh | lbfgs | 0.0001 | 0.0001 |
| 0 | 0.717391 | 0.911232 | 20 | identity | sgd | 0.001 | 0.001 |
| 173 | 0.711957 | 0.909420 | 100 | identity | adam | 1e-06 | 1e-06 |
| 318 | 0.711957 | 0.887681 | 100 | logistic | sgd | 1e-06 | 0.0001 |
| 714 | 0.711957 | 0.876812 | 100 | relu | adam | 1e-07 | 1e-06 |
| 402 | 0.711957 | 1.000000 | 20 | tanh | lbfgs | 1e-05 | 1e-05 |
| 593 | 0.711957 | 0.813406 | 20 | relu | adam | 1e-06 | 1e-06 |
| 637 | 0.711957 | 0.934783 | (20, 20) | relu | adam | 1e-05 | 1e-05 |
| 244 | 0.706522 | 0.864130 | (20, 20) | logistic | sgd | 1e-07 | 0.001 |
| 414 | 0.706522 | 0.820652 | 20 | tanh | adam | 1e-07 | 1e-06 |
| 533 | 0.706522 | 0.891304 | 100 | tanh | adam | 1e-06 | 1e-06 |
| 638 | 0.701087 | 0.969203 | (20, 20) | relu | adam | 1e-06 | 1e-05 |
| 120 | 0.701087 | 0.922101 | 100 | identity | sgd | 0.001 | 0.001 |
| 219 | 0.701087 | 0.925725 | 20 | logistic | adam | 1e-07 | 1e-05 |
| 243 | 0.695652 | 0.782609 | (20, 20) | logistic | sgd | 1e-06 | 0.001 |
| 222 | 0.695652 | 1.000000 | 20 | logistic | lbfgs | 1e-05 | 1e-05 |
| 52 | 0.695652 | 0.813406 | 20 | identity | adam | 1e-05 | 1e-06 |
| 199 | 0.695652 | 0.800725 | 20 | logistic | sgd | 1e-07 | 0.0001 |
| 712 | 0.695652 | 0.887681 | 100 | relu | adam | 1e-05 | 1e-06 |
| 415 | 0.690217 | 0.983696 | 20 | tanh | lbfgs | 0.001 | 1e-06 |
| 191 | 0.690217 | 1.000000 | 20 | logistic | lbfgs | 0.0001 | 0.001 |
| 190 | 0.690217 | 0.998188 | 20 | logistic | lbfgs | 0.001 | 0.001 |
| 1 | 0.690217 | 0.998188 | 20 | identity | sgd | 0.0001 | 0.001 |

## 7.6 Grid Search whole route + Extra data in the corner

Below you can find the GridSearchCV results for the whole training route + the extra training data in the selected corner of the route. The models have been placed in descending order according to their *mean_ test score* (mean validation score).

| | mean_test_score | mean_train_score | param_hidden_layer_sizes | param_activation | param_solver | param_tol | param_learning_rate_init |
|---|---|---|---|---|---|---|---|
| 403 | 0.821429 | 1.000000 | 100 | tanh | lbfgs | 0.0001 | 1e-05 |
| 81 | 0.812500 | 0.950893 | (100, 100) | identity | sgd | 0.001 | 0.001 |
| 0 | 0.808036 | 0.912202 | 20 | identity | sgd | 0.001 | 0.001 |
| 212 | 0.808036 | 0.992560 | (100, 100) | logistic | adam | 1e-05 | 1e-05 |
| 367 | 0.803571 | 0.997024 | (20, 20) | tanh | lbfgs | 0.0001 | 0.0001 |
| 131 | 0.803571 | 0.937500 | 20 | logistic | adam | 1e-05 | 1e-05 |
| 233 | 0.799107 | 1.000000 | 20 | relu | lbfgs | 1e-05 | 0.0001 |
| 358 | 0.799107 | 0.994048 | (20, 20) | tanh | lbfgs | 0.0001 | 0.001 |
| 114 | 0.794643 | 0.986607 | 20 | logistic | lbfgs | 0.001 | 0.001 |
| 404 | 0.794643 | 1.000000 | 100 | tanh | lbfgs | 1e-05 | 1e-05 |
| 348 | 0.794643 | 0.994048 | 20 | tanh | lbfgs | 0.001 | 1e-05 |
| 260 | 0.794643 | 1.000000 | (20, 20) | relu | lbfgs | 1e-05 | 0.0001 |
| 54 | 0.794643 | 0.910714 | 100 | identity | sgd | 0.001 | 0.001 |
| 412 | 0.794643 | 1.000000 | (100, 100) | tanh | lbfgs | 0.0001 | 0.001 |
| 125 | 0.790179 | 1.000000 | 20 | logistic | lbfgs | 1e-05 | 0.0001 |
| 375 | 0.790179 | 0.991071 | (20, 20) | tanh | lbfgs | 0.001 | 1e-05 |
| 365 | 0.790179 | 1.000000 | (20, 20) | tanh | adam | 1e-05 | 0.0001 |
| 357 | 0.790179 | 0.991071 | (20, 20) | tanh | lbfgs | 0.001 | 0.001 |
| 133 | 0.790179 | 1.000000 | 20 | logistic | lbfgs | 0.0001 | 1e-05 |
| 186 | 0.790179 | 1.000000 | 100 | logistic | lbfgs | 0.001 | 1e-05 |
| 88 | 0.785714 | 1.000000 | (100, 100) | identity | lbfgs | 0.0001 | 0.001 |
| 82 | 0.785714 | 1.000000 | (100, 100) | identity | sgd | 0.0001 | 0.001 |
| 61 | 0.785714 | 1.000000 | 100 | identity | lbfgs | 0.0001 | 0.001 |
| 177 | 0.785714 | 1.000000 | 100 | logistic | lbfgs | 0.001 | 0.0001 |
| 251 | 0.785714 | 1.000000 | (20, 20) | relu | lbfgs | 1e-05 | 0.001 |
| 296 | 0.785714 | 1.000000 | 100 | relu | lbfgs | 1e-05 | 1e-05 |
| 312 | 0.785714 | 1.000000 | (100, 100) | relu | lbfgs | 0.001 | 0.0001 |
| 411 | 0.785714 | 1.000000 | (100, 100) | tanh | lbfgs | 0.001 | 0.001 |
| 323 | 0.785714 | 1.000000 | (100, 100) | relu | lbfgs | 1e-05 | 1e-05 |
| 223 | 0.781250 | 1.000000 | 20 | relu | lbfgs | 0.0001 | 0.001 |
| ... | ... | ... | ... | ... | ... | ... | ... |